\documentclass[letterpaper]{article} 
\usepackage[preprint]{aaai2027}  
\usepackage[hyphens]{url}  
\usepackage{graphicx} 
\usepackage{natbib}  
\usepackage{caption} 
\usepackage{algorithm}
\usepackage{algorithmic}
\usepackage{amsmath}
\usepackage{amssymb}
\usepackage{booktabs}
\usepackage{multirow}
\usepackage{array}

\usepackage{newfloat}
\usepackage{listings}
\DeclareCaptionStyle{ruled}{labelfont=normalfont,labelsep=colon,strut=off} 
\floatstyle{ruled}
\newfloat{listing}{tb}{lst}{}
\floatname{listing}{Listing}

\usepackage{booktabs}

\title{Self-Consistent Neural Control: Learning Trajectories for Implicit Equilibrium}

\author{
    Hengyi Zhou\textsuperscript{\rm 1},
        Haobo Fang\textsuperscript{\rm 3},
    Xiaonan Huang\textsuperscript{\rm 1},
    Dezhong Tong\textsuperscript{\rm 1},
    M. Khalid Jawed\textsuperscript{\rm 2}
}

\affiliations{
    \textsuperscript{\rm 1}University of Michigan, Ann Arbor, MI, USA\\
    \textsuperscript{\rm 2}University of California, Los Angeles, CA, USA\\
        \textsuperscript{\rm 3}Carnegie Mellon University, Pittsburgh, PA, USA\\
    hengyizh@umich.edu
}

\begin{document}

\maketitle

\begin{abstract}
One big challenge in physical AI is interacting with systems whose states are defined by implicit equilibrium. Learning in these systems is difficult because their responses can be multistable, path-dependent, and sensitive over long horizons. Adjoint learning provides an efficient way to optimize trajectories without unrolling iterative equilibrium solvers, but its gradient relies on the frozen-tangent approximation, which neglects higher-order sensitivity terms and can become inaccurate when the local equilibrium varies rapidly. In this work, we introduce Self-Consistent Neural Control (SCNC), optimizing trajectories along which sensitivity variations are explicitly controlled. We derive a mathematical bound on the resulting gradient error and develop a practical finite-difference approximation. We validate the framework on deformable objects, which are representative of implicit equilibrium systems, ranging from simple quadratic energy models such as elastic rods to complex systems including thin-walled tubes, measuring tapes, and Slinky-like structures. Compared with the adjoint learning and the gradient-free iCEM baselines, SCNC achieves faster convergence and more stable optimization by reducing switching to undesirable equilibrium branches, doubling the stable optimization rate over adjoint learning (83.3\% vs. 41.7\%) and reducing the average loss by more than 1.5 orders of magnitude on thin-walled tubes, where sensitivity variations are pronounced.
\end{abstract}


\section{Introduction}

Physical AI aims to enable robots to perceive, reason, and interact with the physical world.
For robots that manipulate compliant objects and structures, many interactions can be modeled through mechanical equilibrium rather than explicit state-transition equations.
The system state is therefore obtained by solving an equilibrium problem subject to control-dependent boundary conditions. 
Efficient learning and control through such implicit models are important for complex robotic interactions and have attracted growing interest \citep{yin2021modeling,tong2026neuralcontrol}.
%

\begin{figure}[t]
    \centering
    \includegraphics[width=\columnwidth]{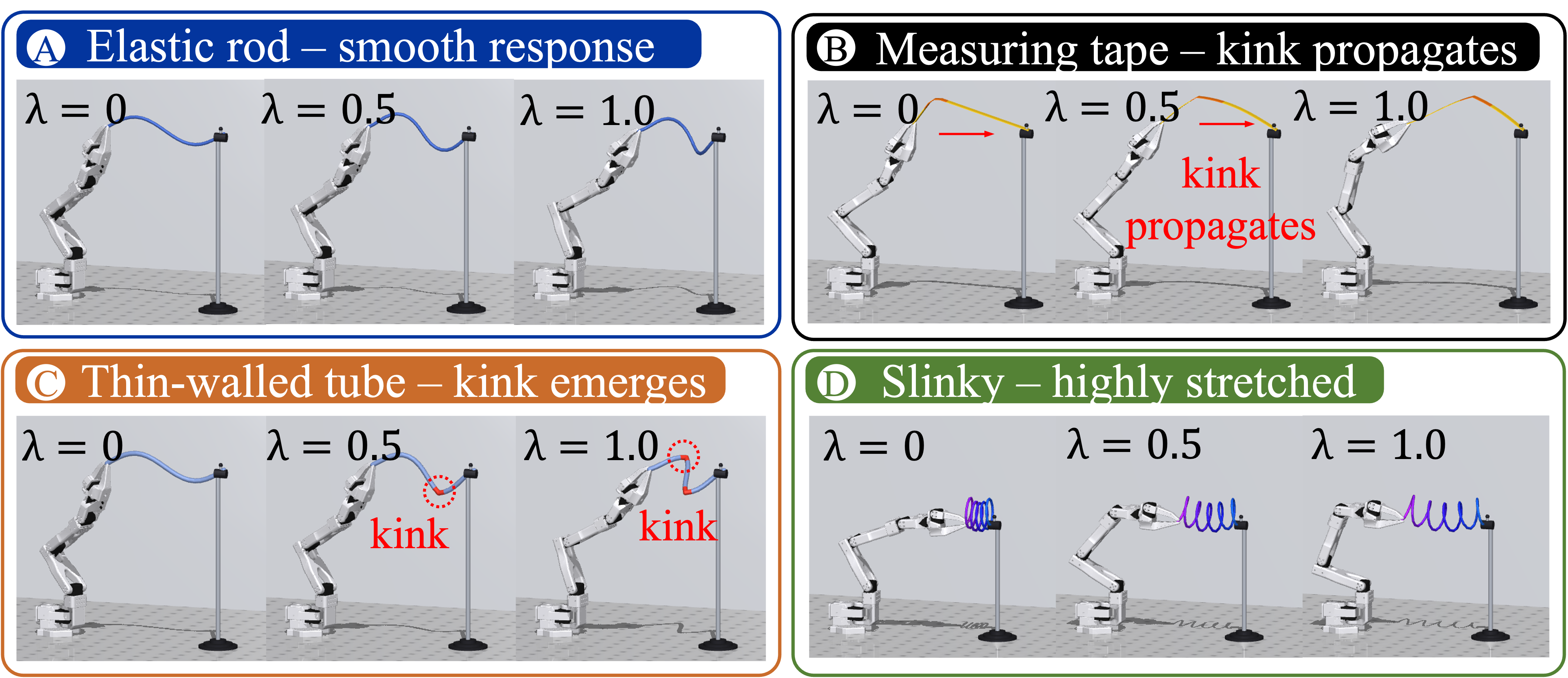}
    \caption{Representative implicit equilibrium systems with increasingly complex local responses. (A) Smooth bending-dominated deformation of an elastic rod. (B) Kink propagation during measuring-tape manipulation. (C) Kink formation in a thin-walled tube. (D) Large stretching and geometric reconfiguration of a Slinky-like structure. $\lambda$ parameterizes the control trajectory.}
    \label{teaser}
\end{figure}

Despite this progress, learning and long-horizon control through implicit equilibrium systems remain challenging, particularly when the equilibrium response varies strongly with the control input.
Multiple stable equilibria may coexist, and the equilibrium branch reached by the system can depend on its deformation history \citep{tong2026neuralcontrol}.
Deformable-object manipulation provides representative examples.
As shown in Fig.~\ref{teaser}, elastic rods provide a relatively well-behaved, bending-dominated case, although the attained equilibrium branch can still depend on the deformation path. 
Measuring tapes and thin-walled tubes are more instability-prone: increasing curvature can induce cross-sectional deformation, kinking, and abrupt equilibrium changes through the Brazier effect~\citep{pastrana2025mechanics} (Fig.~\ref{teaser}(B)--(C)). 
More complex structures, such as Slinky-like systems, exhibit strongly coupled bending, stretching, and large geometric reconfiguration (Fig.~\ref{teaser}(D)).
Although these systems differ mechanically, they illustrate a more general computational challenge: the local sensitivity of an implicit state to control can vary rapidly along the optimization trajectory.

Existing approaches differentiate such systems by unrolling iterative solvers, differentiating explicit learned surrogates, or applying implicit differentiation to converged equilibria or fixed points \citep{hu2020difftaichi,sanchezgonzalez2020learning,bai2019deep}. 
Explicit surrogates provide direct gradients but require representative training data and may accumulate model error over long horizons \citep{sanchezgonzalez2020learning,janner2019mbpo}. 
Implicit models are commonly evaluated through iterative solvers such as Newton--CG or fixed-point iteration.
Differentiating through these iterations requires unrolling and backpropagating through the full solution process~\citep{williams1990efficient}, resulting in deeply nested computational graphs and substantial computational and memory costs.
These limitations motivate methods that differentiate directly through the converged implicit condition without unrolling the internal solver iterations.


Recently, adjoint learning was proposed to retain the implicit forward solve while avoiding costly solver unrolling \citep{tong2026neuralcontrol}. 
Rather than differentiating through the internal iterations of the equilibrium solver, it constructs proxy dynamics whose gradients are determined by the sensitivity of the converged equilibrium state to controllable quantities. 
To improve computational efficiency over long horizons, adjoint learning further adopts a frozen-tangent approximation, in which higher-order variations of the implicit-state sensitivity along the trajectory are neglected~\citep{tong2026neuralcontrol}.

The accuracy of this approximation, however, relies on the implicit-state sensitivity varying sufficiently slowly along the optimization trajectory. This condition can fail in strongly nonlinear regimes, where small changes in control may produce large variations in the local sensitivity or drive the system toward different solution branches \citep{mroz1994sensitivity,geng2021training}. The neglected higher-order sensitivity variations can then introduce substantial gradient errors, producing poor search directions and driving optimization toward highly sensitive or undesirable regions of the implicit solution landscape. This exposes a fundamental inconsistency: the gradient approximation assumes locally smooth implicit-state sensitivity, yet the optimization trajectory itself is unconstrained from entering regions where this assumption becomes invalid.

\begin{figure}[t]
    \centering
    \includegraphics[width=\columnwidth]{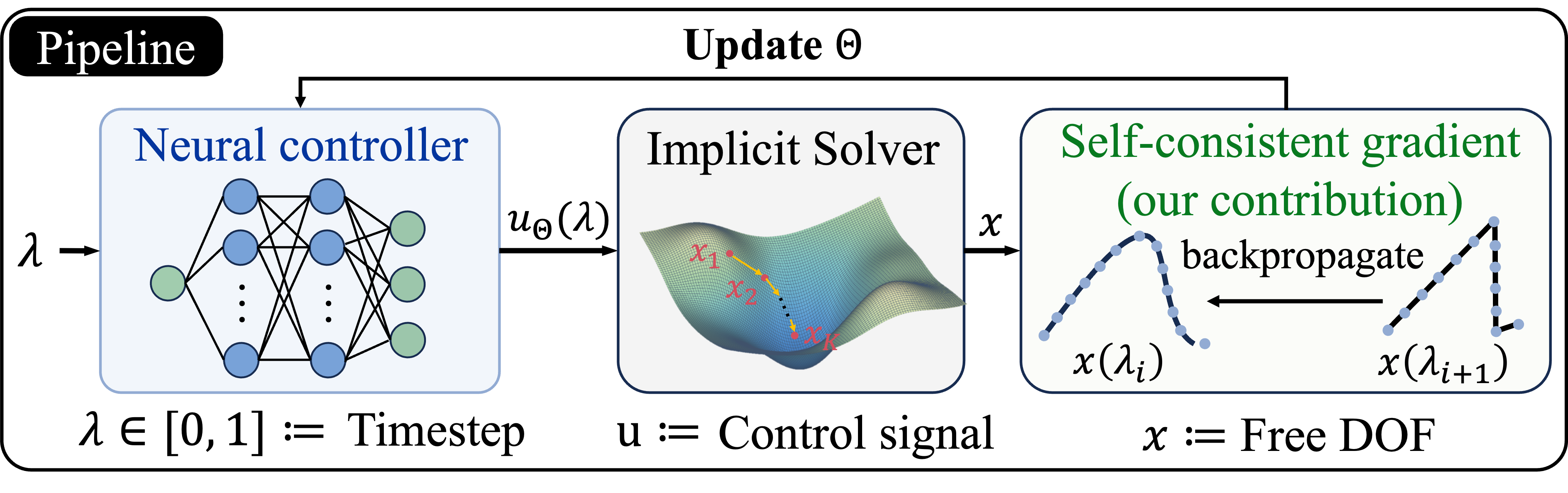}
    \caption{Pipeline for the learning process. The policy generates control signals, and the implicit model determines the corresponding equilibrium states. The task objective is evaluated on the resulting trajectory, and gradients are propagated to optimize the policy. SCNC regularizes the learned trajectory to maintain the reliability of the gradient approximation.}
    \label{pipeline}
\end{figure}

We address this inconsistency with Self-Consistent Neural Control (SCNC), a framework for learning control policies in implicit equilibrium systems with both smooth and rapidly varying equilibrium sensitivities (Fig.~\ref{pipeline}). SCNC is self-consistent in the sense that it regularizes the optimization trajectory to satisfy the local smoothness condition on which its own gradient approximation relies. Rather than assuming that higher-order sensitivity variations are negligible, SCNC explicitly controls their magnitude along the trajectory. We derive a bound relating these variations to the resulting gradient error, making gradient validity a consequence of the optimization process rather than an external assumption. We further develop a practical finite-difference formulation that estimates these sensitivity variations using neighboring converged implicit states to avoid high computational cost.

We validate SCNC across a range of implicit equilibrium systems, including elastic rods, thin-walled tubes, measuring tapes, and Slinky-like structures with neural-network-based energy models. The benchmarks span both smooth bending-dominated responses and strongly instability-prone regimes, with validation in simulation and real robotic manipulation experiments. Compared with adjoint learning and the gradient-free iCEM baseline, SCNC achieves lower task loss and higher optimization success rates while retaining computational efficiency comparable to adjoint learning.

In summary, the main contributions of this work are:
\begin{itemize}
\item We characterize theoretically and numerically how rapidly varying implicit-state sensitivities can introduce substantial errors under the frozen-tangent approximation, leading to unreliable gradients along the optimization trajectory.

\item We develop Self-Consistent Neural Control (SCNC), derive a bound relating higher-order implicit-state sensitivity variations to gradient error, and introduce a practical finite-difference formulation that estimates and regularizes these variations along the optimization trajectory.

\item We validate SCNC across smooth and instability-prone implicit equilibrium systems, including elastic rods, thin-walled tubes, measuring tapes, and Slinky-like structures, demonstrating improved task performance and optimization robustness over adjoint learning and the gradient-free iCEM baseline in both simulation and real-world experiments.
\end{itemize}

    
    

\section{Related Work}

\subsection{Learning and Long-Horizon Control of Implicit Equilibrium Systems}

Learning-based control can rely on either explicit dynamics models or implicitly defined system responses. Explicit learned dynamics and differentiable simulators enable gradient-based trajectory optimization and physical rollouts \citep{hu2020difftaichi,sanchezgonzalez2020learning}, but long-horizon optimization can suffer from accumulated model error \citep{janner2019mbpo}.

For quasi-static systems, control instead induces sequences of nonlinear equilibria. Examples include energy-based shell and rod models \citep{grinspun2003discreteshells,bergou2008discreteelasticrods} and learned fixed-point models such as deep equilibrium networks \citep{bai2019deep}. With coexisting equilibria, warm-started solves can become path- and branch-dependent \citep{bretl2014quasistatic,tong2026neuralcontrol}. Adjoint learning differentiates such sequences without unrolling the inner solver, but its gradient can become unreliable when local sensitivities vary rapidly \citep{tong2026neuralcontrol}, which motivates SCNC.

\subsection{Differentiation and Gradient Reliability Through Implicit Equilibria}

Gradients through equilibria can be obtained by unrolling the numerical solver or differentiating the converged optimality condition. Unrolling is general but incurs horizon-dependent memory cost and truncated-solve errors \citep{domke2012generic,gruslys2016memory}. Implicit differentiation instead solves a linear adjoint system, as used in optimization layers, deep equilibrium networks, and modular frameworks \citep{amos2017optnet,bai2019deep,blondel2022efficient}, but remains sensitive to Jacobian conditioning and solve accuracy \citep{geng2021training}. Error guarantees for approximate implicit or one-step differentiation typically rely on contraction or local-convergence assumptions \citep{grazzi2020iteration,bolte2023onestep}.

In nonlinear mechanics, tangent stiffness can become singular near critical equilibria and continuation may switch branches, requiring specialized sensitivity treatments \citep{mroz1994sensitivity,parente2008sensitivity}. Adjoint learning avoids solver unrolling with a frozen-tangent proxy, but does not control sensitivity variation along instability-prone paths \citep{tong2026neuralcontrol}. SCNC instead treats gradient reliability as trajectory-dependent, regularizing the learned trajectory so that sensitivity variation remains controlled and gradient validity becomes an outcome of optimization.

\subsection{Continuation and Branch Selection near Instabilities}

Quasi-static manipulation can be viewed as planning over a manifold of equilibrium configurations \citep{bretl2014quasistatic}, while direct parameter continuation fails at simple folds \citep{dickson2007condition}. Arc-length and pseudo-arclength methods track response curves through limit points and snap-through \citep{riks1979incremental,crisfield1981fast,dickson2007condition}; near bifurcations, critical-point detection and branch-switching methods access secondary branches \citep{fujii1992branch,uecker2014pde2path}, while deflation reveals additional coexisting solutions \citep{farrell2015deflation}.

These methods primarily improve equilibrium-branch tracking. SCNC instead optimizes a task-driven control trajectory while regulating the sensitivity variation that governs gradient reliability. Thus, continuation targets forward branch tracking, whereas SCNC targets reliable gradient-based trajectory learning.

\section{Methodology}

\subsection{Adjoint learning method}
Consider an implicit equilibrium system whose free equilibrium state is denoted as $\mathbf x$ and the controllable part is denoted as $\mathbf z$. $\lambda$ is used to denote the continuation parameter. $\mathbf x$ and $\mathbf z$ are functions of $\lambda$. We will use $\mathbf x(\lambda)$ or $\mathbf z(\lambda)$ when we try to indicate the function itself, but $\mathbf x_k$ or $\mathbf z_k$ for the specific states at $\lambda = k$. 

In a normal explicit system, state transitions can be easily expressed in the form of $\mathbf x_{k+1} = \mathcal{T}(\mathbf x_k,\mathbf z_k)$. However, when it comes to implicit systems, we have
\begin{equation}
    \mathcal{G}(\mathbf x(\lambda),\mathbf z(\lambda))
    =
    0.
\end{equation}

Consider the setting in Fig.~\ref{teaser}(A), $\mathbf z$ represents the controllable DOF state, which is the state of the rod grabbed by the grippers, $\mathbf x$ represents the free DOF state, and $\mathcal{G}$ is the implicit equilibrium. The robot should learn how to control $\mathbf z$ to push $\mathbf x$ to the target $\mathbf x^*$ following the rule of $\mathcal{G}$.

In adjoint learning \citep{tong2026neuralcontrol}, a neural controller parameterized by $\Theta$ takes $\lambda$ as input and generates a sequence of control signals $u_\Theta(\lambda)$. The controller drives $z$ through known dynamics $f(z(\lambda),u(\lambda))$, while the corresponding free state $x$ is obtained from the implicit equilibrium condition. Since each equilibrium may require many internal solver iterations, directly backpropagating through all solver steps becomes prohibitively expensive over a long horizon.

Sensitivity is defined as:
\begin{equation}
    \mathbf S(\lambda)
    :=
    \frac{\partial \mathbf x}{\partial \mathbf z}(\lambda),
    \qquad
    \mathbf G_x(\lambda) \mathbf S(\lambda)
    =
    - \mathbf G_z(\lambda),
\end{equation}
where $\mathbf{G}_x=\nabla_x\mathcal{G}(\mathbf{x},\mathbf{z})$ and $\mathbf{G}_z=\nabla_z\mathcal{G}(\mathbf{x},\mathbf{z})$. In specific dynamics, $\mathcal{G}$ is often defined as $\mathcal{G}(\mathbf x, \mathbf z) = \nabla_xE(\mathbf x, \mathbf z) - F^{ext}$ and thus $\mathbf G_x$ and $\mathbf G_z$ become tangent stiffness and mixed stiffness respectively. The sensitivity $\mathbf S (\lambda)$ is never formed explicitly; instead, products $\mathbf{S}^\top v$ are computed via single adjoint linear solve. 

The proxy dynamics described by sensitivity along the homotopy path are:
\begin{equation}
    \label{eq:proxy}
    \frac{d \mathbf{x}}{d \lambda}
    =
    \mathbf{S}(\lambda) \frac{d \mathbf{z}}{d \lambda}
    =
    \mathbf{S}(\lambda) f(\mathbf{z}(\lambda), \mathbf{u}(\lambda)),
\end{equation}
and the gradient of the loss with respect to the neural network parameters follows from Eq.~(\ref{eq:proxy}):
\begin{equation}
    \label{eq:gradient} 
    \frac{d \mathcal{L}}{d \Theta}
    =
    \int_0^1 (\frac{\partial \mathbf{u}_\Theta(\lambda)}{\partial\Theta})^\top \frac{d \mathcal{L}}{d \mathbf{u}(\lambda)} d\lambda,
\end{equation}
\begin{equation}
    \label{eq:dL_du} 
    \frac{d \mathcal{L}}{d \mathbf{u}(\lambda)}
    =
    \mathbf{B}(\lambda)^\top (g(\lambda)
    +
    \mathbf{S}(\lambda)^\top \mathbf{a}(\lambda)).
\end{equation}

where $a(\lambda) = \frac{\partial \mathcal{L}}{\partial x}(\lambda)$ and $g(\lambda) = \frac{\partial \mathcal{L}}{\partial z}(\lambda)$ are proxy system costates. $\mathbf A(\lambda) = \frac{\partial f}{\partial \mathbf{z}}(z(\lambda), u(\lambda))$ and $\mathbf B(\lambda) = \frac{\partial f}{\partial \mathbf{u}}(z(\lambda), u(\lambda))$ are two constants for fixed task describing how boundary DOF are updated by control signals. Deduction details of Eq.~(\ref{eq:proxy}) (\ref{eq:gradient}) (\ref{eq:dL_du}) can be found in technical supplement section 1 or adjoint learning paper \citep{tong2026neuralcontrol}.

\subsection{Error from frozen-tangent approximation}
Eq.~(\ref{eq:proxy}) (\ref{eq:gradient}) (\ref{eq:dL_du}) and the transition of costate rely on a frozen-tangent approximation: the equilibrium sensitivity $S(\lambda)$ is treated as locally constant during the backward pass. This amounts to neglecting the variation of $S(\lambda)$ along the homotopy trajectory:
\begin{equation}
    \label{eq:ds_dlambda}
    \frac{d \mathbf S}{d \lambda}
    =
    - \mathbf G_x^{-1}(\frac{d\mathbf G_x}{d\lambda} \mathbf{S}
    +
    \frac{d \mathbf G_z}{d\lambda}),
\end{equation}
where
\begin{equation}
    \label{eq: dGx}
    \frac{d\mathbf G_x}{d\lambda}
    =
    \nabla^3_{xxx}E \cdot \frac{d\mathbf{x}}{d\lambda}
    +
    \nabla^3_{xxz}E \cdot \frac{d\mathbf{z}}{d\lambda},
\end{equation}
and
\begin{equation}
    \label{eq: dGz}
    \frac{d \mathbf G_z}{d\lambda}
    =
    \nabla^3_{xzx}E \cdot \frac{d\mathbf{x}}{d\lambda}
    +
    \nabla^3_{xzz}E \cdot \frac{d\mathbf{z}}{d\lambda},
\end{equation}
The frozen-tangent approximation claims that for the system with smoothly varying equilibrium -- such as elastic rods with simple quadratic energy -- these higher-order variations are often small. Thus these terms can be neglected to avoid calculating third-order derivatives of the energy functional which is really expensive.

However, for instability-prone systems like Slinky-like structures, measuring tapes and thin-walled tubes, both $G_x(\lambda)$ and $G_z(\lambda)$ can vary rapidly along the homotopy path. In these regimes, the neglected higher-order terms in sensitivity can be fatal. Numerical details for detecting $ \frac{d \mathbf{S}}{d \lambda} $ are included in technical supplement section 7.

\subsection{Self-consistent learning}
Our goal is to constrain the gradient error coming from the frozen-tangent approximation in an acceptable range without any expensive computation. 

Note that throughout this paper, $\| \cdot \|$ denotes the Euclidean norm when applied to vectors and the Frobenius norm when applied to matrices or tensors.

Define:
\begin{equation}
    \epsilon
    :=
    \sup_{\lambda \in [0,1]}(\| G_x(\lambda)^{-1} \frac{dG_x}{d\lambda}\|
    +
    \| G_x(\lambda)^{-1} \frac{dG_z}{d\lambda}\|)
\end{equation}

The quantity $\epsilon$ measures the inverse-weighted stiffness
variation along the realized homotopy trajectory. We claim that this
quantity controls the error introduced by the frozen-tangent
approximation up to condition-number constants.

\textbf{Theorem 1.}
Assume that $\mathbf{G}_x(\lambda)$ is nonsingular and that
$\mathbf{S}(\lambda)$, $\mathbf{A}(\lambda)$,
$\mathbf{a}(\lambda)$, and
$\mathbf{B}(\lambda)
\frac{\partial \mathbf{u}_{\Theta}(\lambda)}{\partial\Theta}$
are uniformly bounded. Then there exists a constant $C>0$,
independent of $\epsilon$, such that
\begin{equation}
    \label{eq:theorem1}
    \left\|
    \nabla_{\Theta}\mathcal{L}_{p}
    -
    \nabla_{\Theta}\mathcal{L}_{e}
    \right\|
    \leq
    C\epsilon.
\end{equation}

\noindent
\textit{Proof.}
Define
\begin{equation}
    \mathbf{K}(\lambda)
    :=
    \mathbf{B}(\lambda) \frac{\partial\mathbf{u}_{\Theta}(\lambda)} {\partial\Theta}.
\end{equation}
Based on Eq.~(\ref{eq:gradient}) (\ref{eq:dL_du}), the exact and proxy gradients (gradient in adjoint learning method) are
\begin{equation}
    \nabla_{\Theta}\mathcal{L}_{i} =
    \int_0^1
    \mathbf{K}^{\top}
    \left(
    \mathbf{g}_{i}
    +
    \mathbf{S}^{\top}\mathbf{a}_{i}
    \right)
    d\lambda,
\end{equation}
where $i=e$ denotes exact gradient, $i=p$ denotes proxy gradient, $\bar{\mathbf{S}}$ denotes the locally frozen sensitivity and $a_e(\lambda) = a_p(\lambda) = \mathbf{a}$ (see in technical supplement section 2). Note that $\mathbf{K}$ is determined by computation graph of neural network itself and the known dynamics $f$. So $\mathbf{K}$ has no difference between exact and proxy gradient.
Therefore,
\begin{equation}
\begin{array}{rcl}
    \nabla_{\Theta}\mathcal{L}_{p}
    -
    \nabla_{\Theta}\mathcal{L}_{e}
    &=&
    \displaystyle
    \int_0^1
    \mathbf{K}^{\top}
    \left[
    \mathbf{g}_{p}
    -
    \mathbf{g}_{e}
    \right.
    \\[1mm]
    &&\displaystyle\left.
    +
    \left(
    \bar{\mathbf{S}}
    -
    \mathbf{S}
    \right)^{\top}
    \mathbf{a}
    \right]
    d\lambda.
\end{array}
\end{equation}

Since uniformly bounded, we can assume $\left\|\mathbf{I}(\lambda)\right\|\leq M_I, I \in \{\mathbf{a}, K, S\}$. Then
\begin{equation}
\begin{aligned}
    \left\|
    \nabla_{\Theta}\mathcal{L}_{p}
    -
    \nabla_{\Theta}\mathcal{L}_{e}
    \right\|
    &\leq
    M_K
    \left(
    \sup_{\lambda\in[0,1]}
    \left\|
    \mathbf{g}_{p}(\lambda)
    -
    \mathbf{g}_{e}(\lambda)
    \right\|
    \right.
    \\[1mm]
    &\left.
    +
    M_a
    \sup_{\lambda\in[0,1]}
    \left\|
    \bar{\mathbf{S}}(\lambda)
    -
    \mathbf{S}(\lambda)
    \right\|
    \right).
\end{aligned}
\label{eq:gradient_error_decomposition}
\end{equation}

From Eq.~(\ref{eq:ds_dlambda}),
\begin{equation}
    \left\|
    \frac{d\mathbf{S}}{d\lambda}
    \right\|
    \leq
    C_S\epsilon,
    \qquad
    C_S:=\max\{M_S,1\}.
\end{equation}
Then
\begin{equation}
    \label{eq:sensitivity_error}
    \sup_{\lambda\in[0,1]} \left\|
    \bar{\mathbf{S}}-\mathbf{S}
    \right\|
    \leq
    \Delta\lambda C_S\epsilon.
\end{equation}
where $\Delta\lambda:=\lambda_{k+1}-\lambda_k$.
The transition equation of the costate $\mathbf g$ and $ \sup_{\lambda\in[0,1]} \left\|\mathbf{A}(\lambda)\right\| \leq M_A$, combined with
Gronwall's inequality give
\begin{equation}
    \left\|
    \mathbf{g}_{p}-\mathbf{g}_{e}
    \right\|
    \leq
    C_g \epsilon.
\end{equation}
Details of evaluation on $\mathbf g$ are included in technical supplement section 3.

Therefore, Eq.~(\ref{eq:theorem1}) holds with
\begin{equation}
    C
    :=
    M_K ( C_g  + M_a C_S \Delta\lambda),
\end{equation}
where $C$ is a constant independent of $\epsilon$.

The premises of theorem 1 are encouraged by the penalty below.

This bound motivates a self-consistent control framework: the controller is optimized not only to minimize the task objective, but also to select trajectories along which the approximation used to compute its gradient remains valid.

\subsection{Finite-difference surrogate penalty}
Theorem 1 suggests constraining the sensitivity variation through
\begin{equation}
    \label{eq:regularization_loss}
    \mathcal{L}_{reg}
    =
    \mu_x
    \int_0^1
    \left\|
    \mathbf{G}_x^{-1}
    \frac{d\mathbf{G}_x}{d\lambda}
    \right\|^2
    d\lambda
    +
    \mu_z
    \int_0^1
    \left\|
    \mathbf{G}_x^{-1}
    \frac{d\mathbf{G}_z}{d\lambda}
    \right\|^2
    d\lambda,
\end{equation}
where $\mu_x$ and $\mu_z$ are regularization weights. However,
directly evaluating and differentiating Eq.~(\ref{eq:regularization_loss})
is expensive because of the derivative and inverse matrix based on Eq. ~(\ref{eq: dGx}) (\ref{eq: dGz}).

We therefore construct a finite-difference approximation. Discretize
the homotopy path into $K$ intervals with $\Delta\lambda=\frac{1}{K}$.
The stiffness derivatives can be approximated by
\begin{equation}
    \frac{d\mathbf{G}_x}{d\lambda}
    \bigg|_{\lambda_k}
    =
    \frac{
    \mathbf{G}_x(\lambda_{k+1})
    -
    \mathbf{G}_x(\lambda_k)}
    {\Delta\lambda}
    +
    \mathcal{O}(\Delta\lambda),
\end{equation}
\begin{equation}
    \frac{d\mathbf{G}_z}{d\lambda}
    \bigg|_{\lambda_k}
    =
    \frac{
    \mathbf{G}_z(\lambda_{k+1})
    -
    \mathbf{G}_z(\lambda_k)}
    {\Delta\lambda}
    +
    \mathcal{O}(\Delta\lambda).
\end{equation}

For any matrix $\mathbf{Q}$,
\begin{equation}
    \frac{\|\mathbf{Q}\|}
    {\sigma_{\max}(\mathbf{G}_x)}
    \leq
    \|\mathbf{G}_x^{-1}\mathbf{Q}\|
    \leq
    \frac{\|\mathbf{Q}\|}
    {\sigma_{\min}(\mathbf{G}_x)},
\end{equation}
where ($\sigma_{\min}$) and ($\sigma_{\max}$) denote the smallest and largest singular values, respectively. Therefore, as long as $\mathbf G_x$ is not strictly singular, the inverse-weighted quantities in Eq.~(22) are controlled. Actually, it is in practical applications and is not only dependent on the configurations but also on Young's modulus. Across all 24 experimental cases under SCNC, the smallest minimum singular value of $\mathbf G_x$ observed over all steps is $3.7\times10^{-4}$, occurring in the thin-walled tube scenario. At this scale, the effect associated with the near-singularity can still be effectively controlled by the $\mu$. We combined the finite-difference method in Eqs.~(23)--(24), which motivates the following condition-number-dependent practical surrogate:
\begin{equation}
    \label{eq:practical_regularization}
    \begin{array}{rcl}
    \mathcal{L}_{reg}^{practical}
    &=&
    \mu_x
    \sum_{k=0}^{K-1}
    \frac{
    \left\|
    \mathbf{G}_x(\lambda_{k+1})
    -
    \mathbf{G}_x(\lambda_k)
    \right\|^2}
    {\Delta\lambda^2}
    \\[2mm]
    &&+
    \mu_z
    \sum_{k=0}^{K-1}
    \frac{
    \left\|
    \mathbf{G}_z(\lambda_{k+1})
    -
    \mathbf{G}_z(\lambda_k)
    \right\|^2}
    {\Delta\lambda^2}.
    \end{array}
\end{equation}

The final training objective is
\begin{equation}
    \mathcal{L}_{total}
    =
    \mathcal{L}_{task}
    +
    \mathcal{L}_{reg}^{practical}.
\end{equation}

Unlike Eq.~(\ref{eq:regularization_loss}),
Eq.~(\ref{eq:practical_regularization}) doesn't contains a matrix
inverse. It only penalizes the variation of the tangent and mixed stiffness matrices
between consecutive equilibrium states, providing a practical
approximation of the original regularization loss. 

Pseudocode is included in technical supplement section 8.

\section{Result}
We evaluate SCNC using a set of 2D deformable manipulation testbeds spanning smooth and instability-prone implicit equilibrium regimes. The experiments include elastic rods, thin-walled tubes, and measuring tapes in both simulation and the real world, together with simulated Slinky-like structures governed by a learned neural energy model.

We use python jax library to realize our Newton iterative simulator combined with line search when minimizing the energy for implicit equilibrium.

We compare our SCNC with adjoint learning and iCEM. All the methods share the same forward rollout method described above but utilize different ways to calculate gradients.

\begin{figure}[t]
    \centering
    \includegraphics[width=\columnwidth]{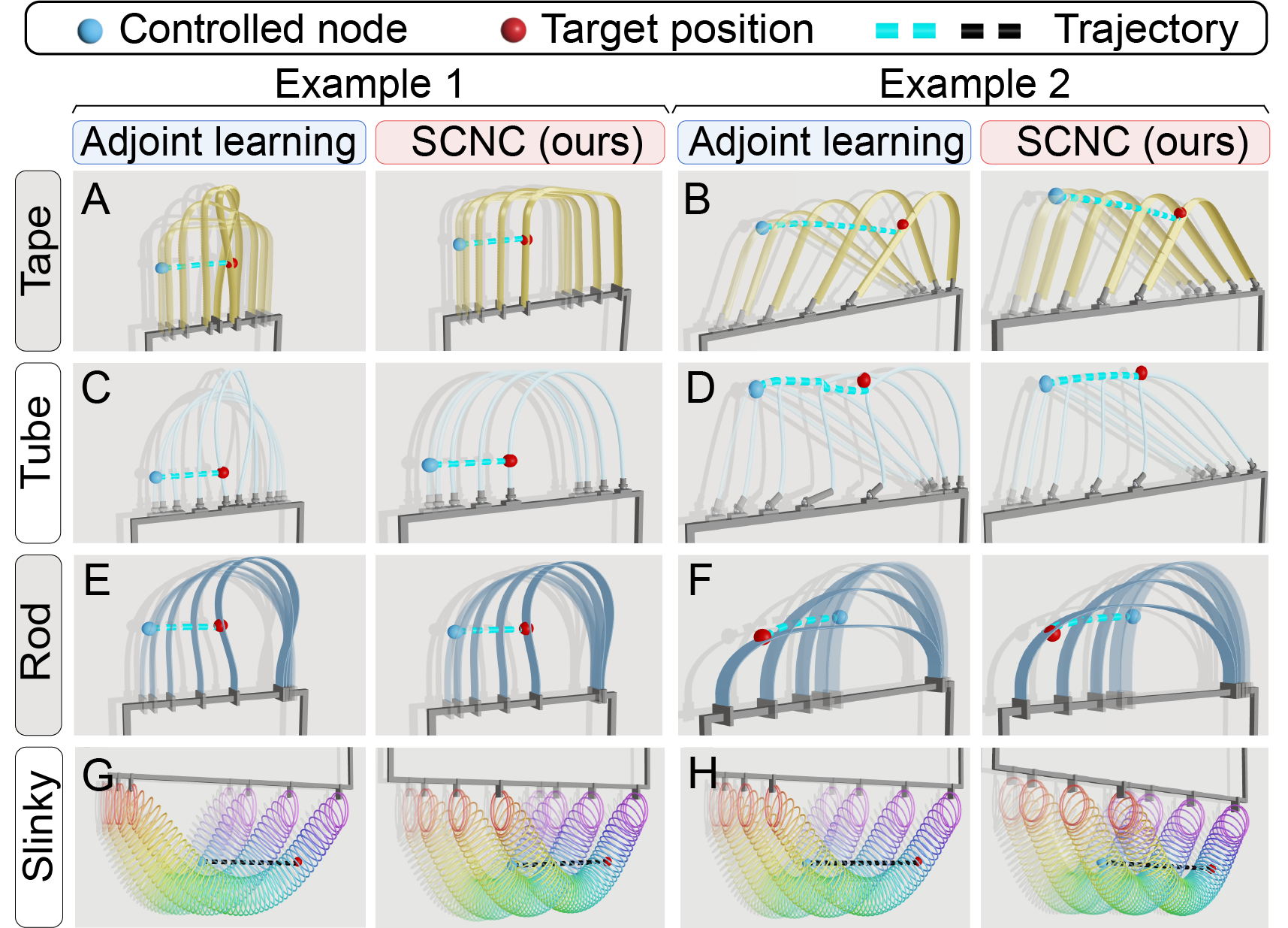}
    \caption{Partial numerical result. Two cases for each object. The boundary conditions are controlled to drive the user-specified nodes to targets. Cases from tape, tube and Slinky shows the stability of SCNC than adjoint learning.}
    \label{render}
\end{figure}

\subsection{Experimental setup}
We model the elastic rods, thin-walled tubes and measuring tapes as center lines with stretching and bending energy. The center lines are discretized into 101 nodes with each node including 2 DOF: $x$ and $y$. For Slinky-like structures, we follow the Slinky model of \citep{li2023rapidly} whose energy is governed by a learned neural network. We choose one with 50 coils and each coils has 3 DOF: $x$, $y$ and $\alpha$ which denotes the orientation of the coil. For all of these, the robot arm will grab the both ends. Models' details are provided in the technical supplement sections 4 and 5.
\subsection{Tasks and metrics}
\label{Tasks and metrics}
For all objects, we use the same node-targeting task, in which the robot manipulates both ends of the object to move a designated node to a target position. The control signal $u_\Theta(\lambda_i)$ varies from $\mathbb{R}^2$ to $\mathbb{R}^4$, including $x$ axis displacement for two ends and the in-plane rotation for two clamps.

Denote the coordinates of chosen node as $\mathbf n$ which is a function of $\lambda$ and target position as $\mathbf n^*$. The task loss is defined as:
\begin{equation}
    \mathcal{L}_{task} = \frac{1}{2} \| \mathbf n(1) - \mathbf n^*\|^2.
\end{equation}

Besides, we define a optimization is stable if the equilibrium doesn't switch to the bad branch -- kink doesn't propagate (change the relative position on the object), disappear and no new kink occurs for tapes and tubes; stiffness of Slinky-like structures doesn't vary too fast (the final distance between left end and right end is less than 1.5 times its original distance).

Each object contains six evaluation cases for every method.

\subsection{Simulation results}
\label{numerical results}
We reported some representative simulation results in Fig.~\ref{render}. 

The rendered figure shows the stability of our method. As described before, our method is less prone to switching to undesirable equilibrium branches. For case A, adjoint learning gives a bad final equilibrium, where the two initial kinks merge into one and the whole tape twists. While SCNC translates the tape as a whole and preserves the stability of the kink when trying to move the controlled point to the target. For case B, adjoint learning propagates a kink initially near the left end toward the right end of the tape while SCNC keeps the relative position of the kink. Similar things happen on tube cases. For cases C and D, adjoint learning's trajectories have new kink occurring while SCNC maintains the original configurations and realizes the target at the same time.

When it comes to the Slinky-like structures, SCNC still shows advantages. For case G, adjoint learning primarily moves only the right end of the Slinky, resulting in a substantial increase in its length. This significantly changes the stiffness and sensitivity of the system, making control much more challenging. In contrast, SCNC again learns to translate the entire Slinky toward the target while avoiding excessive stretching. Similar behaviors can be observed in case H.

For elastic rods (cases E and F), the policies learned by adjoint learning and SCNC are nearly indistinguishable, which is consistent with our expectations. Owing to the simple quadratic energy form, elastic rods exhibit smooth and well-behaved sensitivity variations. As a result, the frozen-tangent approximation underlying adjoint learning remains sufficiently accurate, allowing it to achieve performance comparable to that of SCNC. 

\subsection{Baselines comparison}
In addition to adjoint learning, we choose iCEM as another baseline because it is widely used for trajectory optimization. \citep{pinneri2021icem}. We select population size $P = 10$, which already imposes a substantial computational burden. All other parameters needed are included in technical supplement section 6. 

\subsubsection{Convergence}
Fig.~\ref{results}(A) shows the score (log$_{10}$(best loss)) vs. epoch loss history. Each thin curve corresponds to the average loss trajectory of a method for one object, whereas the thick curve represents the overall mean loss trajectory.
The results show that both adjoint learning and SCNC substantially outperform iCEM, while SCNC exceeds adjoint learning by 0.601 points in score and achieves an approximately 4$\times$ lower average loss at epoch 30.

The largest performance gap between SCNC and adjoint learning is observed for thin-walled tubes, where the average loss differs by more than 1.5 orders of magnitude. Technical Supplement section 7 further shows that this system exhibits the largest sensitivity variations, together with the largest difference in sensitivity variation between the two methods, supporting the effectiveness of explicitly controlling sensitivity variations in instability-prone systems.

In contrast, for elastic rods with relatively small sensitivity variations, adjoint learning already performs well. Consequently, the gap between SCNC and adjoint learning is much smaller, as expected.

Statistical significance analysis is included in technical supplement section 9.

\subsubsection{Stability}
\begin{figure}[t]
    \centering
    \includegraphics[width=\columnwidth]{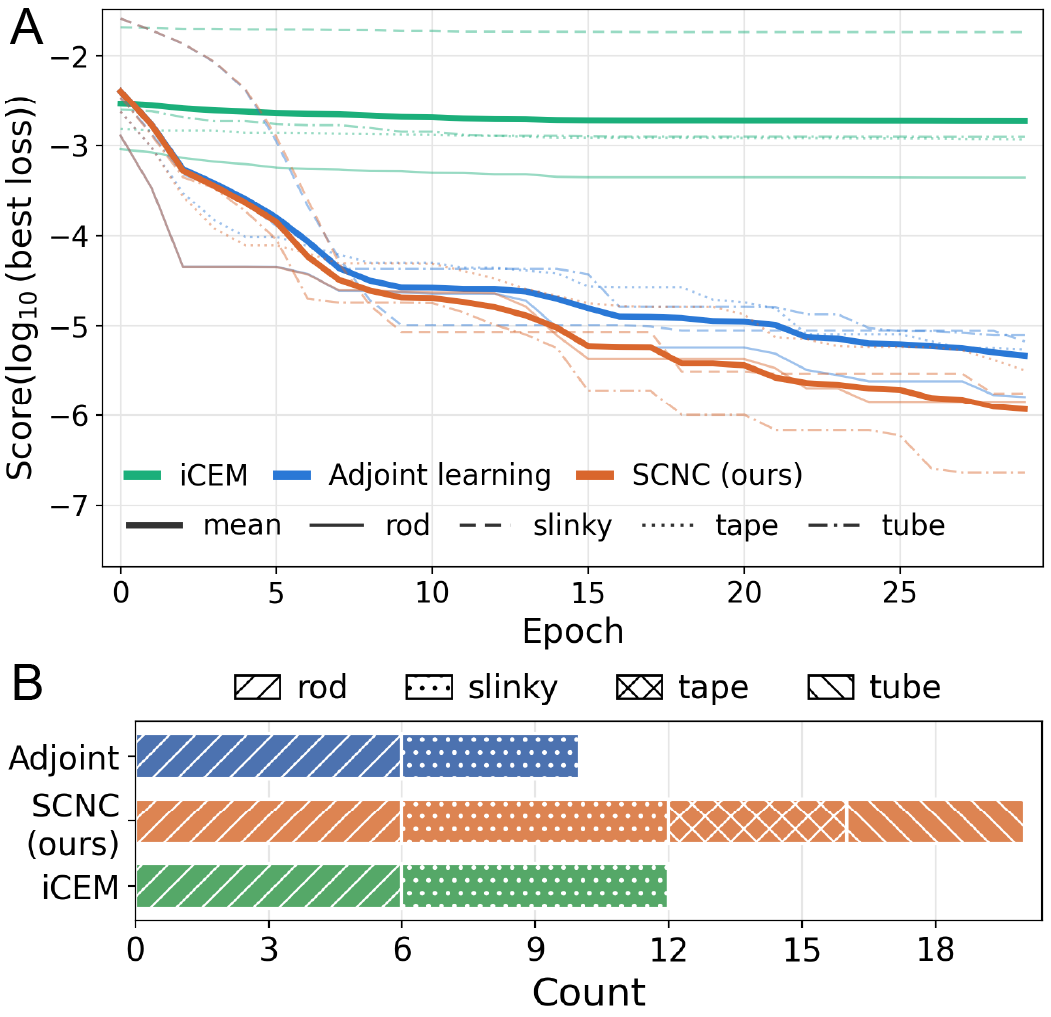}
    \caption{(A) Loss score vs. epoch for different methods and objects. (B) Stable optimization count for different methods and objects.}
    \label{results}
\end{figure}

Fig.~\ref{results}(B) compares the stability of each method across different objects according to the stable optimization criterion defined in Section~\ref{Tasks and metrics}. Among the 24 evaluation cases, SCNC achieves stable optimization in 20 cases (83.3\%), substantially outperforming iCEM with 12 cases (50.0\%) and adjoint learning with 10 cases (41.7\%). In particular, SCNC remains stable on several instability-prone tape and tube tasks where both baselines fail, demonstrating its improved robustness to abrupt equilibrium transitions and rapidly varying sensitivities.

\subsubsection{Computational cost}
As Table~\ref{computational cost} reports, SCNC incurs almost no additional overhead compared with adjoint learning, with the average time per epoch increasing by only about 8\%. Nevertheless, it remains substantially more efficient than iCEM. This demonstrates the practicality of our finite-difference surrogate, which significantly improves optimization performance while introducing only a marginal computational overhead.

Peak memory consumption is comparable across all three methods because the libraries occupy the majority of memory needed. Finally, the computational cost for the Slinky-like structure is reported separately in the Technical Supplement section 6.4 because its energy is represented by a neural network.

\begin{table*}[t]
    \centering
    \setlength{\tabcolsep}{5.5pt}
    \renewcommand{\arraystretch}{1.08}

    \begin{tabular}{
        l
        cc
        cc
        cc
        cc
        cc
    }
        \toprule
        \multirow{2}{*}{\textbf{Method}}
        & \multicolumn{2}{c}{\textbf{Tape}}
        & \multicolumn{2}{c}{\textbf{Tube}}
        & \multicolumn{2}{c}{\textbf{Rod}}
        & \multicolumn{2}{c}{\textbf{Average}}
        \\
        \cmidrule(lr){2-3}
        \cmidrule(lr){4-5}
        \cmidrule(lr){6-7}
        \cmidrule(lr){8-9}
        \cmidrule(l){10-11}

        & Time[s]
        & Mem.[MB]
        & Time[s]
        & Mem.[MB]
        & Time[s]
        & Mem.[MB]
        & Time[s]
        & Mem.[MB]
        \\
        \midrule

        Adjoint learning
        & 11.56 & 1834.8
        & 2.85 & 1865.9
        & 2.00 & 1680.3
        & 5.47 & 1793.7
        \\

        SCNC(ours)
        & 11.77 & 1849.0
        & 3.26 & 1873.5
        & 2.76 & 1762.5
        & 5.93 & 1828.3
        \\

        iCEM
        & 113.08 & 1846.4
        & 21.40 & 1871.0
        & 17.09 & 1761.5
        & 50.52 & 1826.3
        \\

        \bottomrule
    \end{tabular}
    \caption{Computational efficiency comparison across different physical systems.
    Each system contains six node-targeting tasks.
    Time denotes time per epoch, and Mem. denotes peak memory usage.}
    \label{computational cost}
\end{table*}

\subsection{Real world experiments}
\begin{figure*}[t]
    \centering
    \includegraphics[width=0.8\textwidth]{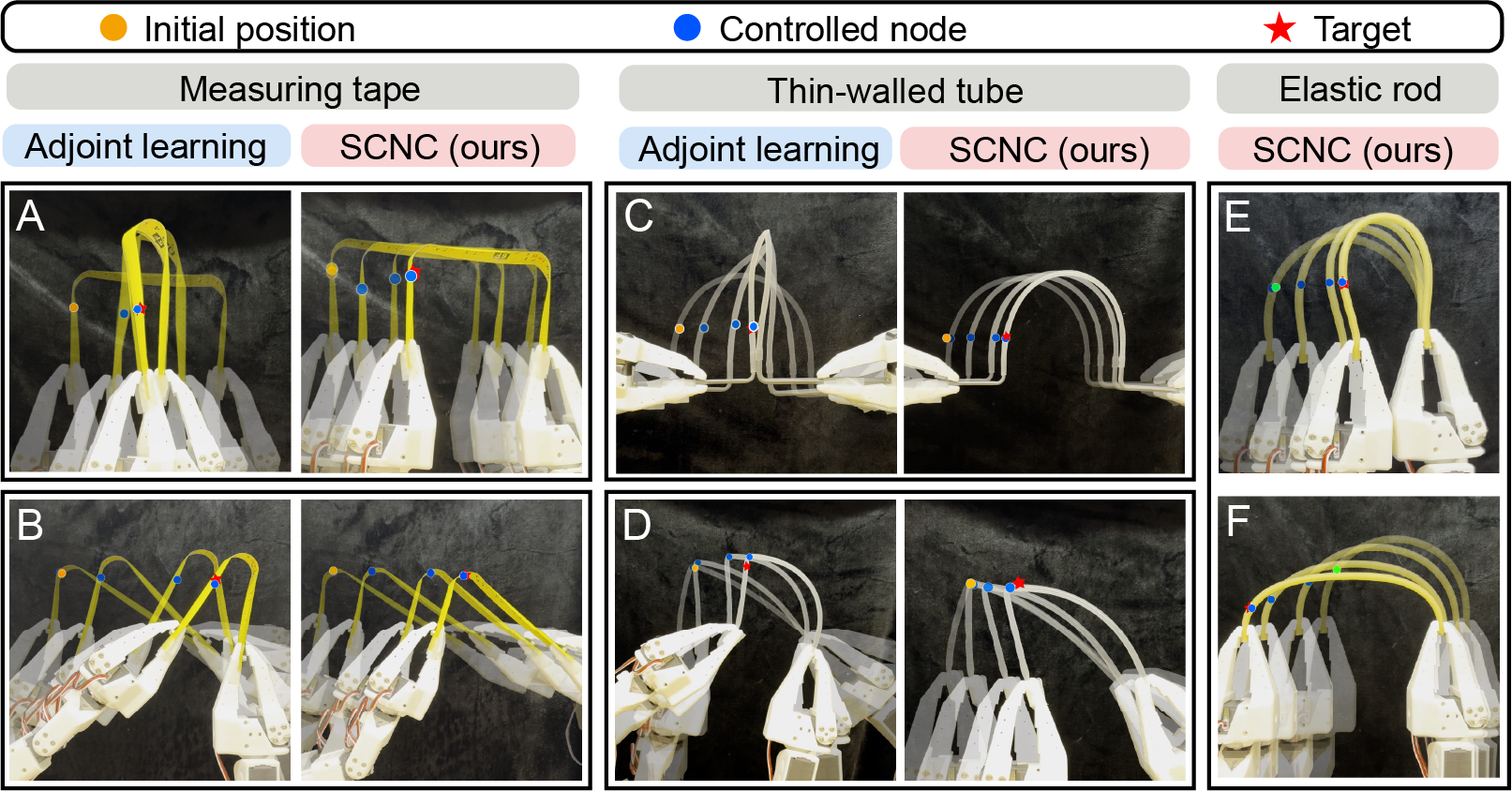}
    \caption{Real world experiments. Opacity increases over time.}
    \label{real world}
\end{figure*}

As shown in Fig.~\ref{real world}, the real-world experiments follow the simulation setup in Fig.~\ref{render}, except that the Slinky-like structure is excluded and only SCNC results are shown for the elastic rod, where SCNC and adjoint learning perform comparably. Matching subfigure labels denote the same experimental cases.

Consistent with Section~\ref{numerical results}, the real-world experiments show that SCNC substantially outperforms adjoint learning on thin-walled tubes and measuring tapes, where instability-induced transitions are more pronounced, while achieving comparable performance on elastic rods. This confirms that the advantages of SCNC extend to real-world robotic manipulation.

Additionally, a complete video of each real-world experimental case is provided in the media supplement, using the same label as in Fig.~\ref{render} and Fig.~\ref{real world}.

\section{Discussion}
Our results suggest that optimization stability in implicit equilibrium systems is closely tied to the validity of the gradient approximation. When local equilibrium sensitivities vary rapidly, the frozen-tangent approximation can produce inaccurate search directions, leading to unstable optimization and degraded task performance. Rather than recovering the neglected higher-order derivatives, SCNC steers optimization toward trajectories where the approximation remains reliable, resulting in more stable learning and improved performance across instability-prone regimes.

Although demonstrated on energy-based mechanics systems, the proposed framework is not restricted to any particular constitutive model. The derivation only assumes an implicitly defined equilibrium together with differentiable local equilibrium sensitivities, making the idea potentially applicable to a broader class of implicit physical systems. The experiments on neural-network energy models provide an indication of this generality.

The current work nevertheless has several limitations. First, although the proposed theory is generally applicable to implicit equilibrium systems, the experimental evaluation is currently limited to node-targeting manipulation tasks. While these tasks capture the optimization challenges caused by rapidly varying equilibrium sensitivities, they do not fully represent the diversity of practical robotic manipulation problems. Future work will evaluate SCNC on more challenging scenarios, including shape control, 3D manipulation and contact-rich interaction. Second, the regularization weights are currently selected empirically based in the Young's modulus of the material (See technical supplement section 7). Developing adaptive regularization strategies that automatically adjust the penalty according to the sensitivity variation is another promising direction for future research.

\section{Conclusion}
We presented Self-Consistent Neural Control (SCNC), a learning framework for implicit equilibrium systems that improves optimization by explicitly controlling the variation of local equilibrium sensitivity. Unlike adjoint learning, which applies the frozen-tangent approximation with high risk, SCNC regularizes the trajectory to maintain the reliability of the approximation. We further derived a theoretical bound relating sensitivity variation to gradient error and proposed an efficient practical finite-difference surrogate. Experimental results on both physical and learned equilibrium models demonstrate that SCNC achieves more stable optimization and better task performance than existing gradient-based and gradient-free baselines, particularly in instability-prone regimes. We hope this work provides a useful perspective for scalable learning and control in implicit physical systems and inspires future research on robust optimization for physical AI.

\bibliography{aaai2027}


\end{document}


\maketitle

\section{Supplemental proof for adjoint learning}
\label{proof: adjoint}
This section derives the proxy dynamics, the costate equations, and the gradient of adjoint learning \citep{tong2026neuralcontrol}, i.e., Eqs.~(4)--(6) of the main paper, without differentiating through the inner equilibrium solver.

\subsection{Implicit sensitivity}
At any converged equilibrium $(\mathbf x(\lambda), \mathbf z(\lambda))$ satisfying $\mathcal G(\mathbf x, \mathbf z) = 0$, assume the realized equilibrium branch is locally smooth and $\mathbf G_x = \nabla_x \mathcal G$ is nonsingular. Differentiating the equilibrium condition with respect to $\mathbf z$ gives
\begin{equation}
    \mathbf G_x\, \frac{\partial \mathbf x}{\partial \mathbf z}
    +
    \mathbf G_z
    =
    0
    \quad\Longrightarrow\quad
    \mathbf G_x \mathbf S(\lambda) = -\mathbf G_z,
\end{equation}
which is the sensitivity definition in the main paper. $\mathbf S$ is never formed explicitly: for any vector $\mathbf v$, the product $\mathbf S^\top \mathbf v$ is obtained by one adjoint linear solve,
\begin{equation}
    \mathbf G_x^\top \mathbf p = \mathbf v,
    \qquad
    \mathbf S^\top \mathbf v = -\mathbf G_z^\top \mathbf p.
\end{equation}

Along the realized branch, the total derivative of $\mathbf x$ with respect to $\lambda$ is
\begin{equation}
    \frac{d\mathbf x}{d\lambda}
    =
    \mathbf S(\lambda)\, \frac{d\mathbf z}{d\lambda}
    +
    \frac{\partial \mathbf x}{\partial \lambda}.
\end{equation}
In our setting $\mathcal G$ has no explicit $\lambda$-dependence (any scheduling is absorbed into $\mathbf z(\lambda)$), so $\partial \mathbf x / \partial \lambda = 0$. Substituting the continuation dynamics $d\mathbf z / d\lambda = f(\mathbf z, \mathbf u)$ yields the proxy dynamics, Eq.~(4) of the main paper:
\begin{equation}
    \label{supp:proxy}
    \frac{d\mathbf x}{d\lambda}
    =
    \mathbf S(\lambda)\, f(\mathbf z(\lambda), \mathbf u(\lambda)).
\end{equation}
Eq.~(\ref{supp:proxy}) is used only for the backward sensitivity propagation; the forward rollout always solves $\mathcal G = 0$ to convergence with warm starts.

\subsection{Discrete adjoint recursion}
Discretize $\lambda \in [0,1]$ into $\{\lambda_k\}_{k=0}^{K}$ with $\Delta\lambda = 1/K$. The forward rollout is
\begin{equation}
    \mathbf u_k = \mathbf u_\Theta(\lambda_k),
    \qquad
    \mathbf z_{k+1} = \mathbf z_k + \Delta\lambda\, f(\mathbf z_k, \mathbf u_k),
\end{equation}
with $\mathbf x_k$ the converged equilibrium at $\mathbf z_k$. Consider the terminal loss $\mathcal L = \phi(\mathbf x_K, \mathbf z_K)$ (the node-targeting loss of the main paper is of this form) and define the costates
\begin{equation}
    \mathbf a_k := \frac{\partial \mathcal L}{\partial \mathbf x_k},
    \qquad
    \mathbf g_k := \frac{\partial \mathcal L}{\partial \mathbf z_k},
\end{equation}
with terminal conditions $\mathbf a_K = \nabla_x \phi$, $\mathbf g_K = \nabla_z \phi$.

The frozen-tangent approximation evaluates $\mathbf S_k$ on the realized rollout and treats it as locally constant during the backward pass, i.e., neglects the derivatives of $\mathbf S_k$ with respect to $\mathbf x_k$, $\mathbf z_k$, and $\lambda_k$. Under this approximation, the Euler discretization of Eq.~(\ref{supp:proxy}), $\mathbf x_{k+1} \approx \mathbf x_k + \Delta\lambda\, \mathbf S_k f(\mathbf z_k, \mathbf u_k)$, has the linearized transitions
\begin{equation}
    \frac{\partial \mathbf z_{k+1}}{\partial \mathbf z_k} = \mathbf I + \Delta\lambda\, \mathbf A_k,
    \qquad
    \frac{\partial \mathbf z_{k+1}}{\partial \mathbf u_k} = \Delta\lambda\, \mathbf B_k,
\end{equation}
\begin{equation}
    \begin{array}{c}
    \displaystyle
    \frac{\partial \mathbf x_{k+1}}{\partial \mathbf x_k} = \mathbf I,
    \qquad
    \frac{\partial \mathbf x_{k+1}}{\partial \mathbf z_k} \approx \Delta\lambda\, \mathbf S_k \mathbf A_k,
    \\[3mm]
    \displaystyle
    \frac{\partial \mathbf x_{k+1}}{\partial \mathbf u_k} \approx \Delta\lambda\, \mathbf S_k \mathbf B_k,
    \end{array}
\end{equation}
where $\mathbf A_k = \frac{\partial f}{\partial \mathbf z}$ and $\mathbf B_k = \frac{\partial f}{\partial \mathbf u}$ are evaluated at $(\mathbf z_k, \mathbf u_k)$. Applying the chain rule backward from $k = K-1$ to $0$,
\begin{equation}
    \mathbf a_k
    =
    \left(\frac{\partial \mathbf x_{k+1}}{\partial \mathbf x_k}\right)^{\!\top} \mathbf a_{k+1}
    =
    \mathbf a_{k+1},
    \label{supp:a-recursion}
\end{equation}
\begin{equation}
    \mathbf g_k
    =
    (\mathbf I + \Delta\lambda\, \mathbf A_k)^\top \mathbf g_{k+1}
    +
    \Delta\lambda\, \mathbf A_k^\top \mathbf S_k^\top \mathbf a_{k+1},
    \label{supp:g-recursion}
\end{equation}
\begin{equation}
    \frac{d\mathcal L}{d\mathbf u_k}
    =
    \Delta\lambda\, \mathbf B_k^\top
    \left(
    \mathbf g_{k+1} + \mathbf S_k^\top \mathbf a_{k+1}
    \right),
    \label{supp:du-recursion}
\end{equation}
where each product $\mathbf S_k^\top(\cdot)$ is computed by the matrix-free adjoint solve above. The gradient follows from the chain rule through the controller,
\begin{equation}
    \frac{d\mathcal L}{d\Theta}
    =
    \sum_{k=0}^{K-1}
    \left(
    \frac{\partial \mathbf u_\Theta(\lambda_k)}{\partial \Theta}
    \right)^{\!\top}
    \frac{d\mathcal L}{d\mathbf u_k}.
    \label{supp:dtheta-discrete}
\end{equation}

\subsection{Continuous limit}
Rearranging Eq.~(\ref{supp:a-recursion}) and Eq.~(\ref{supp:g-recursion}) and letting $\Delta\lambda \to 0$ gives the backward costate ODEs, integrated from $\lambda = 1$ to $0$:
\begin{equation}
    -\frac{d\mathbf a}{d\lambda} = 0,
    \qquad
    -\frac{d\mathbf g}{d\lambda}
    =
    \mathbf A(\lambda)^\top
    \left(
    \mathbf g(\lambda) + \mathbf S(\lambda)^\top \mathbf a(\lambda)
    \right),
    \label{supp:costate-ode}
\end{equation}
with $\mathbf a(1) = \nabla_x \phi$ and $\mathbf g(1) = \nabla_z \phi$. In the same limit, Eq.~(\ref{supp:du-recursion}) and Eq.~(\ref{supp:dtheta-discrete}) become
\begin{equation}
    \frac{d\mathcal L}{d\mathbf u(\lambda)}
    =
    \mathbf B(\lambda)^\top
    \left(
    \mathbf g(\lambda) + \mathbf S(\lambda)^\top \mathbf a(\lambda)
    \right),
\end{equation}
\begin{equation}
    \frac{d\mathcal L}{d\Theta}
    =
    \int_0^1
    \left(
    \frac{\partial \mathbf u_\Theta(\lambda)}{\partial \Theta}
    \right)^{\!\top}
    \frac{d\mathcal L}{d\mathbf u(\lambda)}\,
    d\lambda,
\end{equation}
which are exactly Eqs.~(5)--(6) of the main paper. Note that for the terminal loss, $\mathbf a(\lambda) = \nabla_x \phi$ is constant along the backward pass; this fact is used in Section~\ref{proof: costate a}. The transition equation of $\mathbf g$ in Eq.~(\ref{supp:costate-ode}) is the one referenced in Section~\ref{proof: costate g}: the frozen-tangent (proxy) costate $\mathbf g_p$ solves Eq.~(\ref{supp:costate-ode}) with the frozen sensitivity $\overline{\mathbf S}$, whereas the exact costate $\mathbf g_e$ solves it with the true sensitivity $\mathbf S$.

\section{Proof for the upper bound of costate $\mathbf a$}
\label{proof: costate a}
Since the exact and proxy methods share the same forward trajectory,
they evaluate the loss on the same equilibrium state and control
trajectory. Therefore,
\begin{equation}
{\bf a}_e(\lambda)
=
{\bf a}_p(\lambda)
=
{\bf a}(\lambda),
\end{equation}
for all $\lambda\in[0,1]$.

In particular, $\mathbf a_p = \mathbf a_e = \mathbf a(1)$ for node-targeting tasks. 

\section{Proof for the upper bound of costate $\mathbf g$}
\label{proof: costate g}
Let ${\bf g}_e$ and ${\bf g}_p$ denote the exact and proxy costates, respectively. Subtracting their costate equations gives
\begin{equation}
    -
    \frac{d}{d\lambda}
    \left(
    {\bf g}_p-{\bf g}_e
    \right)
    =
    {\bf A}^\top
    \left(
    {\bf g}_p-{\bf g}_e
    \right)
    +
    {\bf A}^\top
    \left(
    \overline{\bf S}-{\bf S}
    \right)^\top
    {\bf a}.
\end{equation}

By the variation-of-constants formula and the common terminal condition
${\bf g}_p(1)={\bf g}_e(1)$,
\begin{equation}
\begin{array}{rcl}
\left\|
{\bf g}_p(\lambda)-{\bf g}_e(\lambda)
\right\|
&\leq&
M_A
\displaystyle
\int_{\lambda}^{1}
\left\|
{\bf g}_p(s)-{\bf g}_e(s)
\right\|
\,ds
\\[6pt]
&&+
M_A M_a
\displaystyle
\int_{\lambda}^{1}
\left\|
\overline{\bf S}(s)-{\bf S}(s)
\right\|
\,ds.
\end{array}
\end{equation}

Using the sensitivity bound established above, we obtain
\begin{equation}
\left\|
{\bf g}_p(\lambda)-{\bf g}_e(\lambda)
\right\|
\leq
M_A
\int_{\lambda}^{1}
\left\|
{\bf g}_p(s)-{\bf g}_e(s)
\right\|
\,ds
+
C_1\epsilon,
\end{equation}
for a constant $C_1>0$ independent of $\epsilon$.

Applying the backward integral form of Gronwall's inequality yields
\begin{equation}
\sup_{\lambda\in[0,1]}
\left\|
{\bf g}_p(\lambda)-{\bf g}_e(\lambda)
\right\|
\leq
C_g\epsilon,
\end{equation}
where $C_g>0$ is independent of $\epsilon$.

\section{Models for elastic rod, thin-walled tube and measuring tape}
\label{models: rod tube and tape}

\subsection{Discretization}
The elastic rod, the thin-walled tube, and the measuring tape share the same planar representation: the object is modeled by its center line, discretized into $n_v=101$ nodes (i.e., $n_e=100$ segments), each carrying 2 DOF $(x, y)$. The full DOF vector therefore has dimension 202, split into the free part $\mathbf x$ and the controlled part $\mathbf z$ (nodes $\{0, 1, 99, 100\}$; the two nodes at each end form a rigid clamp held by the gripper).

Let $\mathbf t_i$ denote the unit tangent of segment $i$ and $l_i$ its rest length. The signed discrete curvature at an interior node is
\begin{equation}
    \kappa_i
    =
    \frac{1}{\bar l_i}
    \cdot
    \frac{2\, \mathbf t_{i-1} \times \mathbf t_i}
    {1 + \mathbf t_{i-1} \cdot \mathbf t_i},
\end{equation}
where $\bar l_i = \frac{1}{2}(l_{i-1}+l_i)$ is the Voronoi length associated with node $i$. Node masses are lumped, $m_i = \rho A \bar l_i$ with cross-sectional area $A = \pi r^2$, and gravity $\mathbf g = (0, -9.81)\,\mathrm{m/s^2}$ acts on every node.

\subsection{Elastic energy}
The total potential energy is
\begin{equation}
    E(\mathbf x, \mathbf z)
    =
    E_s + E_b + E_g,
\end{equation}
with stretching, bending, and gravitational terms
\begin{equation}
    E_s
    =
    \frac{1}{2}\, EA \sum_{k=1}^{n_e} \varepsilon_k^2\, l_k,
    \qquad
    \varepsilon_k = \frac{\|\mathbf e_k\|}{l_k} - 1,
\end{equation}
\begin{equation}
    E_b
    =
    \frac{1}{2} \sum_{i} \frac{EI_i\,(\kappa_i - \bar\kappa_i)^2}{\bar l_i},
    \qquad
    E_g = -\sum_i m_i\, \mathbf g \cdot \mathbf q_i,
\end{equation}
where $\mathbf e_k$ is the current edge vector, $\bar\kappa_i = 0$ (straight rest state), and $\mathbf q_i$ is the position of node $i$. The stretching stiffness is $EA = E\pi r^2$ and the base bending stiffness is $EI_0 = E\pi r^4/4$ unless specified directly (the measuring tape uses a prescribed $EI_0$; see Section~\ref{parameters}).

\subsection{Curvature-dependent bending stiffness}
The three objects differ only in how the effective bending stiffness $EI_i$ depends on the local curvature. For elastic rods,
\begin{equation}
    EI = EI_0,
\end{equation}
which yields a smooth quadratic bending energy. For thin-walled tubes, the Brazier effect softens the section symmetrically once the curvature magnitude exceeds a critical value $\kappa^*$:
\begin{equation}
    EI = EI_0\left(1 - c\,H(|\kappa| - \kappa^*)\right),
\end{equation}
where $H$ is the Heaviside function, $\kappa^*$ is the critical curvature, and $c\in(0,1)$ is the constant Brazier softening factor. For measuring tapes, the transverse cross-sectional camber makes the response one-sided: bending toward the concave side softens beyond $\kappa^*$, whereas bending toward the convex side stiffens:
\begin{equation}
    EI = EI_0\left(1 - c\,H(\kappa - \kappa^*) + F\,H\!\left(\frac{-\kappa}{d}\right)\right),
\end{equation}
where $F$ is the constant stiffening factor and $d$ is the stiffening ramp width.

In the implementation, the Heaviside functions are replaced by smooth sigmoids so that the energy remains $C^\infty$ and the tangent stiffness stays well defined: the softening gate uses $\sigma\!\left((\kappa^2 - \kappa^{*2})/\varepsilon^2\right)$ with $\varepsilon^2 = w\,\kappa^{*2}$ and smoothing width $w = 0.1$, and the tape stiffening gate uses $\sigma(-\kappa/d)$. For tapes we additionally include a small curvature-gradient regularization energy
\begin{equation}
    E_{grad}
    =
    \frac{1}{2}\, EI_0\, l_g^2 \sum_i \left(\frac{\kappa_{i+1}-\kappa_i}{\bar l_i}\right)^2 \bar l_i,
\end{equation}
where $l_g$ is a regularization length on the order of two edge lengths; this sets a finite fold width and prevents curvature localization into a single vertex.

\subsection{Equilibrium solver and boundary control}
Given the controlled DOF $\mathbf z$, the free equilibrium $\mathbf x$ is obtained by minimizing $E(\mathbf x, \mathbf z)$ with a damped-Newton method implemented in jax: the Hessian is symmetrized and its eigenvalues are floored to keep the search direction a descent direction, followed by Armijo backtracking line search (Armijo constant $10^{-4}$, shrink factor $0.5$). The iteration terminates when the gradient norm falls below $\mathrm{tol}\cdot EI_0/L^2$, where $L$ is the rod length, with at most 300 outer iterations per solve.

The controlled DOF are driven by the control signal through explicit Euler integration along the homotopy parameter,
\begin{equation}
    \mathbf z_{k+1} = \mathbf z_k + \Delta\lambda\, \mathbf B\, \mathbf u_{\Theta}(\lambda_k),
\end{equation}
where $\mathbf B$ maps the control channels to the boundary DOF. All physical parameter values are listed in Section~\ref{parameters}.

\section{Models for Slinky-like structure}
\label{models: slinky}

\subsection{Discretization}
The Slinky-like structure is modeled as a chain of $n_c = 50$ rigid coils. Each coil $i$ carries 3 DOF: the in-plane position $(x_i, y_i)$ of its center and an independent director angle $\alpha_i$ describing the orientation of the coil plane (a Cosserat-type rotational DOF, not derived from the center line), for a total of 150 DOF.

\subsection{Neural network energy}
Unlike the rod, tube, and tape, the Slinky has no analytic constitutive law in our experiments. Its entire elastic potential is represented by a frozen pre-trained Euclidean-symmetric neural network (ESNN) energy following \citet{li2023rapidly}, which we re-implement in jax so that gradients and Hessians of the energy are available to the equilibrium solver and to the adjoint pass.

For every triplet of neighboring coils $(i-1, i, i+1)$, the 9 raw coordinates are reduced to 6 Euclidean invariants by removing rigid-body motion. With $\Delta\mathbf q_1 = (x_i - x_{i-1},\, y_i - y_{i-1})$, $\Delta\mathbf q_2 = (x_{i+1} - x_i,\, y_{i+1} - y_i)$, and the mean director angles $\beta_1 = \frac{1}{2}(\alpha_{i-1} + \alpha_i)$, $\beta_2 = \frac{1}{2}(\alpha_i + \alpha_{i+1})$, the invariants of the first edge are
\begin{equation}
    \xi_1 = \cos\beta_1\, \Delta x_1 + \sin\beta_1\, \Delta y_1,
\end{equation}
\begin{equation}
    \eta_1 = -\sin\beta_1\, \Delta x_1 + \cos\beta_1\, \Delta y_1,
\end{equation}
\begin{equation}
    \delta\alpha_1 = \alpha_i - \alpha_{i-1},
\end{equation}
and analogously $(\xi_2, \eta_2, \delta\alpha_2)$ for the second edge. Each invariant vector is augmented four ways by the two chiral reflections (mirroring about the $x$- and $y$-axes) to enforce chirality equivariance, and the six channels are rescaled by $(10^3, 10^3, 10^1, 10^3, 10^3, 10^1)$ before entering the network.

The network maps the 6 scaled invariants to a scalar triplet energy: a linear layer $6 \to 32$, followed by 5 densely connected blocks of width 32 with Softplus activations ($\beta = 10$) whose outputs are concatenated (final feature width $32 \times 6 = 192$), an element-wise square, and a final linear layer $192 \to 1$. The total elastic energy is the sum over all 48 interior triplets and the 4 chirality augmentations. The network weights are trained as in \citet{li2023rapidly} and kept frozen throughout policy learning; only the control policy is optimized.

\subsection{Equilibrium and boundary control}
The total potential energy is
\begin{equation}
    E(\mathbf x, \mathbf z)
    =
    E_{ESNN}(\mathbf x, \mathbf z)
    -
    \sum_i m\, \mathbf g \cdot \mathbf q_i,
\end{equation}
where gravity acts on the $(x, y)$ DOF of every coil with per-coil mass $m$, and the $\alpha$ DOF carry a rotational inertia $\frac{1}{2} m R^2$ with coil radius $R$. Equilibria are computed with the same damped-Newton solver as in Section~\ref{models: rod tube and tape} (at most 200 outer iterations per solve).

The coils $\{0, 1, 48, 49\}$ are clamped: all 3 DOF of the two coils at each end are fixed and moved rigidly. The control signal is 2-dimensional, $\mathbf u = (\Delta x_L, \Delta x_R)$, translating the left and right clamp horizontally; it is applied by the same Euler integration along $\lambda$ as before. The task is the node-targeting task of the main paper: the loss penalizes the final $(x, y)$ position of one designated coil relative to a target, and $\alpha$ does not enter the loss.

\begin{algorithm}[t]
\caption{SCNC (adjoint learning: $\mu_x = \mu_z = 0$)}
\label{alg:scnc}
\begin{algorithmic}[1]
\REQUIRE policy $\mathbf u_\Theta$; energy $E$; homotopy steps $K$ ($\Delta\lambda = 1/K$); epochs $N$; weights $\mu_x, \mu_z$; target $\mathbf n^*$; bounds $\mathbf u_{max}$
\STATE initialize $\Theta$; Adam optimizer
\FOR{$epoch = 1$ \TO $N$}
    \STATE reset simulator: $\mathbf z_0 \leftarrow$ initial boundary state, $\mathbf x_0 \leftarrow$ initial equilibrium
    \FOR{$k = 0$ \TO $K-1$}
        \STATE $\mathbf u_k \leftarrow \mathbf u_\Theta(\lambda_k) = \mathbf u_{max} \tanh\!\left(0.5\, \mathrm{MLP}_\Theta(\lambda_k)\right)$;\quad $\mathbf z_{k+1} \leftarrow \mathbf z_k + \Delta\lambda\, f(\mathbf z_k, \mathbf u_k)$
        \STATE $\mathbf x_{k+1} \leftarrow \arg\min_{\mathbf x} E(\mathbf x, \mathbf z_{k+1})$ by damped Newton, warm-started from $\mathbf x_k$
        \STATE cache $\mathbf G_{x,k} \leftarrow \nabla^2_{xx} E + \varepsilon_{reg} \mathbf I$,\; $\mathbf G_{z,k} \leftarrow \nabla^2_{xz} E$,\; $\mathbf B_k \leftarrow \partial f/\partial \mathbf u$,\; $\mathbf x_{k+1}$,\; $\mathbf z_{k+1}$
    \ENDFOR
    \STATE $\mathcal L_{task} \leftarrow \frac{1}{2}\|\mathbf n_K - \mathbf n^*\|^2$
    \STATE $\mathcal L_{reg}^{practical} \leftarrow \frac{1}{\Delta\lambda^2} \sum_{k} \big(\mu_x \|\mathbf G_{x,k+1} - \mathbf G_{x,k}\|_F^2$ \\ \hspace{5.2em} $+\, \mu_z \|\mathbf G_{z,k+1} - \mathbf G_{z,k}\|_F^2\big)$
    \STATE $d\mathcal L_{total} / d\Theta \leftarrow$ Algorithm~\ref{alg:adjoint}
    \STATE clip gradient norm to $1.0$; Adam step on $\Theta$
    \STATE snapshot $\Theta$ if $\mathcal L_{task}$ is the best so far
\ENDFOR
\RETURN best-snapshot policy $\Theta$
\end{algorithmic}
\end{algorithm}

\begin{algorithm}[t]
\caption{Self-consistency gradient}
\label{alg:adjoint}
\begin{algorithmic}[1]
\REQUIRE cached $\{\mathbf G_{x,k}, \mathbf G_{z,k}, \mathbf B_k, \mathbf x_{k+1}, \mathbf z_{k+1}\}_{k=0}^{K-1}$ from the rollout of Algorithm~\ref{alg:scnc}
\STATE \textbf{procedure} \textsc{STprod}$(k, \mathbf v)$: solve $\mathbf G_{x,k}^\top \mathbf p = \mathbf v$; \textbf{return} $-\mathbf G_{z,k}^\top \mathbf p$ \COMMENT{$= \mathbf S_k^\top \mathbf v$}
\STATE $\mathbf a \leftarrow \nabla_x \mathcal L_{task}$;\quad $\mathbf g \leftarrow \nabla_z \mathcal L_{task}$
\FOR{$k = K-1$ \TO $0$}
    \STATE $d\mathcal L_{task} / d\mathbf u_k \leftarrow \Delta\lambda\, \big(\mathbf B_{K-1}^\top \mathbf g + \mathbf B_k^\top\, \textsc{STprod}(k, \mathbf a)\big)$
\ENDFOR
\STATE $d\mathcal L_{reg}^{practical} / d\mathbf u_K \leftarrow 0$
\FOR{$k = K-1$ \TO $0$ \COMMENT{skipped when $\mu_x = \mu_z = 0$}}
    \STATE $\mathbf a_{reg} \leftarrow \nabla_{x} \Psi_k$;\quad $\mathbf g_{reg} \leftarrow \nabla_{z} \Psi_k$, \\ where $\Psi_k = \big( \langle \nabla^2_{xx} E,\, \partial \mathcal L_{reg}^{practical} / \partial \mathbf G_{x,k} \rangle$ \\ \hspace{3.2em} $+\, \langle \nabla^2_{xz} E,\, \partial \mathcal L_{reg}^{practical} / \partial \mathbf G_{z,k} \rangle \big)\big|_{(\mathbf x_{k+1}, \mathbf z_{k+1})}$
    \STATE $d\mathcal L_{reg}^{practical} / d\mathbf u_k \leftarrow d\mathcal L_{reg}^{practical} / d\mathbf u_{k+1}$ \\ \hspace{5.2em} $+\, \Delta\lambda\, \mathbf B_k^\top \big( \mathbf g_{reg} + \textsc{STprod}(k, \mathbf a_{reg}) \big)$
\ENDFOR
\RETURN $d\mathcal L_{total} / d\Theta = \sum_{k=0}^{K-1} \left(\partial \mathbf u_\Theta(\lambda_k) / \partial \Theta\right)^\top \big( d\mathcal L_{task} / d\mathbf u_k$ \\ \hspace{5.2em} $+\, d\mathcal L_{reg}^{practical} / d\mathbf u_k \big)$
\end{algorithmic}
\end{algorithm}

\section{Parameters}
\label{parameters}

\subsection{Hyperparameters for learning}
The policy network $\Theta$ is a fully connected MLP that maps the scalar homotopy parameter $\lambda$ to the $m$-dimensional control signal ($m = 2$ to $4$ depending on the task), with two hidden layers of width 64 and ReLU activations. Weights are initialized with Kaiming-uniform initialization ($a = \sqrt 5$) and zero biases. The raw output $\mathbf o$ is squashed to respect per-channel control bounds,
\begin{equation}
    \mathbf u = \mathbf u_{max} \tanh(0.5\, \mathbf o),
\end{equation}
with $\mathbf u_{max}$ set so that the per-step boundary increments satisfy $|\Delta x| \le 0.01$~m and $|\Delta\theta| \le 0.05$~rad (i.e., rate bounds $0.01/\Delta\lambda$ and $0.05/\Delta\lambda$). Both adjoint learning and SCNC share this architecture and all hyperparameters in Table~\ref{tab:learning-hparams}; the only difference is whether the regularization $\mathcal{L}_{reg}^{practical}$ is enabled.

The number of epochs is set to 30 because, across all objects and methods, the training loss essentially plateaus by that point: continuing beyond 30 epochs yields no significant further decrease and only adds computational cost. The number of homotopy steps $K = 30$ is a trade-off between simulator capability and computation time: with fewer steps (larger $\Delta\lambda$) the per-step boundary increments become too large for the warm-started Newton solver to reliably track the current equilibrium branch, whereas more steps increase the per-epoch cost, which grows linearly with $K$, without a noticeable gain in accuracy.

\begin{table}[t]
    \centering
    \setlength{\tabcolsep}{4pt}
    \begin{tabular}{ll}
        \toprule
        Hyperparameter & Value \\
        \midrule
        Network architecture & $1 \to 64 \to 64 \to m$ \\
        Activation & ReLU \\
        Optimizer & Adam \\
        Learning rate & 0.01 \\
        Epochs & 30 \\
        Homotopy steps $K$ & 30 ($\Delta\lambda = 1/30$) \\
        Gradient clipping (norm) & 1.0 \\
        Translation bound per step & 0.01 m \\
        Rotation bound per step & 0.05 rad \\
        Random seed & 42 (all cases and methods) \\
        Precision & float64 \\
        Device for numerical results & Apple M4 on mac OS \\
        Device for computational cost & Intel U9 on Ubuntu 24.04 \\
        \bottomrule
    \end{tabular}
    \caption{Learning hyperparameters shared by adjoint learning and SCNC.}
    \label{tab:learning-hparams}
\end{table}

The regularization weights $\mu_x$ and $\mu_z$ of Eq.~(22) in the main paper are selected empirically per object, scaling inversely with the magnitude of the stiffness entries (and hence with the Young's modulus of the material); their values are given in Table~\ref{tab:reg-weights}. Concretely, $\mu_x$ and $\mu_z$ are tuned so that, under the initial (untrained) policy, the two penalty terms $\mu_x \sum_k \|\Delta\mathbf G_x\|_F^2/\Delta\lambda^2$ and $\mu_z \sum_k \|\Delta\mathbf G_z\|_F^2/\Delta\lambda^2$ are both on the order of $10^{-6}$. This keeps the strength of the regularization comparable across objects despite their vastly different stiffness scales, and small relative to the task loss while still influencing the optimization.

\begin{table}[t]
    \centering
    \begin{tabular}{lcc}
        \toprule
        Object (task) & $\mu_x$ & $\mu_z$ \\
        \midrule
        Rod & $2 \times 10^{-17}$ & $1 \times 10^{-14}$ \\
        Tube & $2 \times 10^{-18}$ & $1 \times 10^{-19}$ \\
        Tape & $3 \times 10^{-26}$ & $4 \times 10^{-25}$ \\
        Slinky-like structure & $1 \times 10^{-9}$ & $1 \times 10^{-7}$ \\
        \bottomrule
    \end{tabular}
    \caption{Regularization weights of SCNC for each object and task type.}
    \label{tab:reg-weights}
\end{table}

\subsection{iCEM}
The iCEM baseline \citep{pinneri2021icem} optimizes the raw control sequence $\mathbf u \in \mathbb{R}^{T \times m}$ directly, clipped element-wise to $[-\mathbf u_{max}, \mathbf u_{max}]$ with the same bounds as the policy network, and runs for the same 30 iterations with the same random seed. Its hyperparameters are listed in Table~\ref{tab:icem-hparams}.

\begin{table}[t]
    \centering
    \begin{tabular}{ll}
        \toprule
        Hyperparameter & Value \\
        \midrule
        Population size $P$ & 10 \\
        Elite fraction & 0.3 (3 elites) \\
        Mean/std smoothing $\alpha$ & 0.25 \\
        Initial std & $0.5\, \mathbf u_{max}$ \\
        Minimum std & $10^{-2}$ \\
        Colored-noise exponent $\beta$ & 2.0 \\
        Elite reuse fraction & 1.0 \\
        Population decay & 1.0 (none) \\
        Best-so-far reinsertion & yes \\
        Iterations & 30 \\
        \bottomrule
    \end{tabular}
    \caption{iCEM hyperparameters.}
    \label{tab:icem-hparams}
\end{table}

\subsection{Physical parameters}
Table~\ref{tab:physical-params} lists the physical and solver parameters of the elastic rod, thin-walled tube, and measuring tape; Table~\ref{tab:slinky-params} lists those of the Slinky-like structure. All objects use $r = 10^{-3}$~m for the center-line cross-sectional radius and gravity $9.81$~$\mathrm{m/s^2}$. For the tape, $EI_0$ is prescribed directly rather than derived from $E\pi r^4/4$, reflecting the strongly anisotropic cross-section; the softening factor is $c = 0.995$, and the curvature-gradient length is $3 \times 10^{-3}$~m.

\begin{table}[t]
    \centering
    \setlength{\tabcolsep}{2pt}
    \small
    \begin{tabular}{lccc}
        \toprule
        Parameter & Rod & Tube & Tape \\
        \midrule
        $E$ [Pa] & $5 \times 10^{6}$ & $2 \times 10^{7}$ & $2 \times 10^{11}$ \\
        $\rho$ [$\mathrm{kg/m^3}$] & $800$ & $1200$ & $7850$ \\
        $EI_0$ [$\mathrm{N \cdot m^2}$] & $3.9 \times 10^{-6}$ & $1.6 \times 10^{-5}$ & $7 \times 10^{-4}$ \\
        $EA$ [N] & $15.7$ & $62.8$ & $6.3 \times 10^{5}$ \\
        Crit.\ curvature $\kappa^*$ [1/m] & --- & $45$ & $10$ \\
        Softening factor $c$ & --- & $0.95$ & $0.995$ \\
        Stiffening factor $F$ & --- & --- & $9$ \\
        Stiffening width $d$ [1/m] & --- & --- & $1$ \\
        Smoothing width $w$ & --- & 0.1 & 0.1 \\
        Newton tol.\ multiplier & 0.1 & 0.1 & 0.1 \\
        Max.\ Newton iterations & 300 & 300 & 300 \\
        \bottomrule
    \end{tabular}
    \caption{Physical and solver parameters for the rod, tube, and tape.}
    \label{tab:physical-params}
\end{table}

\begin{table}[t]
    \centering
    \begin{tabular}{ll}
        \toprule
        Parameter & Value \\
        \midrule
        Number of coils & 50 \\
        Coil radius $R$ & 0.033 m \\
        Coil mass $m$ & $3 \times 10^{-3}$ kg \\
        Rotational inertia $\frac{1}{2} m R^2$ & $1.6 \times 10^{-6}$ $\mathrm{kg \cdot m^2}$ \\
        Initial end-to-end span & 0.186 m \\
        Newton tolerance multiplier & 0.1 \\
        Max Newton iterations & 200 \\
        \bottomrule
    \end{tabular}
    \caption{Physical and solver parameters for the Slinky-like structure.}
    \label{tab:slinky-params}
\end{table}

\subsection{Computational cost for the Slinky-like structure}
Because the Slinky energy is a neural network rather than an analytic constitutive law, its computational cost is reported separately from the main-paper table. Table~\ref{tab:slinky-cost} reports the time per epoch and peak memory averaged over the six Slinky evaluation cases, measured with the same protocol as in the main paper. The relative ordering matches the analytic systems: SCNC adds a moderate overhead over adjoint learning for the finite-difference stiffness penalty (the overhead is larger here because differentiating the penalty requires third-order derivatives of the neural energy), while iCEM remains several times more expensive than both.

\begin{table}[t]
    \centering
    \begin{tabular}{lcc}
        \toprule
        Method & Time/Ep. [s] & Peak Mem. [MB] \\
        \midrule
        Adjoint learning & 5.95 & 1762.0 \\
        SCNC (ours) & 7.46 & 1846.6 \\
        iCEM & 45.02 & 1798.2 \\
        \bottomrule
    \end{tabular}
    \caption{Computational cost on the Slinky-like structure, averaged over six cases.}
    \label{tab:slinky-cost}
\end{table}

\section{Numerical results for the sensitivity variation}
\begin{figure*}[t]
    \centering
    \includegraphics[width=1.0\textwidth]{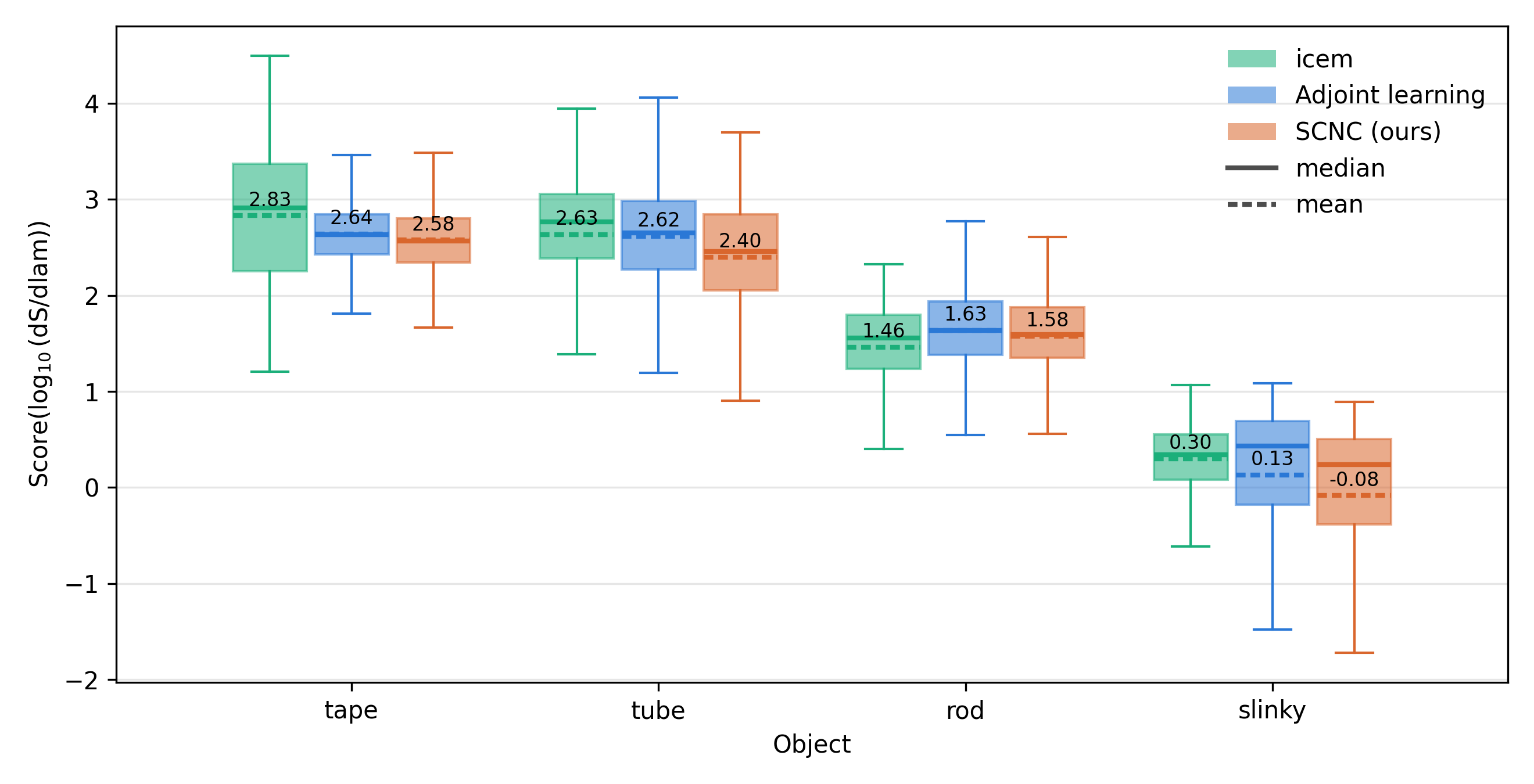}
    \caption{Per-step sensitivity variation along the realized homotopy trajectories, shown as box plots of $\log_{10}\left(\|\mathbf S_{k+1} - \mathbf S_k\|_F / \Delta\lambda\right)$ for each object and method. The instability-prone tape and tube exhibit sensitivity variations one to two orders of magnitude larger than the smooth rod and the Slinky-like structure. SCNC attains the lowest mean variation on most objects, with the largest reduction relative to adjoint learning on the tube, consistent with its largest performance gain in the main paper.}
    \label{ds_dlam}
\end{figure*}

We evaluate the finite-difference variation $\|\mathbf S_{k+1} - \mathbf S_k\|_F / \Delta\lambda$ at every homotopy step along the realized trajectory, where each $\mathbf S_k$ is obtained from the cached tangent stiffness via the products of Algorithm~\ref{alg:adjoint}; no additional solves beyond those already performed during training are required. Figure~\ref{ds_dlam} summarizes the resulting distributions per object and method on a $\log_{10}$ scale.

Two observations support the analysis in the main paper. First, the magnitude of the sensitivity variation separates the smooth systems from the instability-prone ones: the tape and tube exhibit variations one to two orders of magnitude larger than the rod and the Slinky-like structure, which explains why adjoint learning remains accurate on the latter but degrades on the former. Second, SCNC consistently shifts the distribution downward relative to adjoint learning, confirming that the surrogate penalty indeed steers optimization toward trajectories with bounded sensitivity variation. The reduction is largest on the tube, mirroring the nearly two-order-of-magnitude improvement in average loss reported in the main paper and providing direct numerical evidence for the mechanism behind SCNC's performance gains.

\label{numerical ds_dlam}

\section{Algorithm pseudocode}
\label{alg pseudocode}
Algorithm~\ref{alg:adjoint} details the backward propagation. Adjoint learning is recovered exactly by setting $\mu_x = \mu_z = 0$, which disables lines 10 of Algorithm~\ref{alg:scnc}; the two methods share every other line. The iCEM baseline replaces both algorithms by the standard colored-noise cross-entropy loop of \citet{pinneri2021icem} over the raw control sequence, using the same forward rollout and the hyperparameters of Table~\ref{tab:icem-hparams}.

\section{Statistical significance and standard deviation of the loss results}
\label{stats significance}
For every task, each method was run once with the shared random seed of Table~\ref{tab:learning-hparams}, so the comparison across methods is paired per task. For each baseline and each of the 24 tasks (4 objects $\times$ 6 cases) we compute, from the recorded loss histories, the paired difference
\begin{equation}
    \Delta
    =
    \log_{10}\!\left(\mathcal L^{best}_{SCNC}\right)
    -
    \log_{10}\!\left(\mathcal L^{best}_{baseline}\right).
\end{equation}
$\Delta < 0$ means SCNC is better, and $\Delta$ is the $\log_{10}$ of the best loss ratio, hence dimensionless and comparable across objects with vastly different loss scales. Table~\ref{tab:stats-significance} reports the per-object mean and standard deviation of $\Delta$ together with the win rate (fraction of tasks with $\Delta < 0$). The overall row averages the four per-object means so that each object weighs equally, with the standard deviation.

\begin{table}[t]
    \centering
    \setlength{\tabcolsep}{3pt}
    \small
    \begin{tabular}{lccccc}
        \toprule
        & & \multicolumn{2}{c}{vs.\ Adjoint learning} & \multicolumn{2}{c}{vs.\ iCEM} \\
        \cmidrule(lr){3-4} \cmidrule(lr){5-6}
        Object & $n$ & $\Delta$ (mean $\pm$ std) & Wins & $\Delta$ (mean $\pm$ std) & Wins \\
        \midrule
        Rod    & 6  & $-0.06 \pm 0.48$ & 2/6   & $-2.49 \pm 0.96$ & 6/6 \\
        Tube   & 6  & $-1.53 \pm 0.38$ & 6/6   & $-3.73 \pm 1.06$ & 6/6 \\
        Tape   & 6  & $-0.23 \pm 0.88$ & 5/6   & $-2.56 \pm 0.40$ & 6/6 \\
        Slinky & 6  & $-0.58 \pm 0.84$ & 6/6   & $-4.02 \pm 0.86$ & 6/6 \\
        \midrule
        Overall & 24 & $-0.60 \pm 0.66$ & 19/24 & $-3.20 \pm 0.79$ & 24/24 \\
        \bottomrule
    \end{tabular}
    \caption{Paired comparison of $\Delta = \log_{10}$(best loss ratio) of SCNC over each baseline, as mean $\pm$ std and win rate per object. The overall row is the mean $\pm$ std across the four per-object means.}
    \label{tab:stats-significance}
\end{table}

To assess significance without distributional assumptions, we apply a two-sided Wilcoxon signed-rank test to the 24 paired differences of each baseline, computing the $p$-value by a sign-flip permutation test on the signed-rank statistic. Against adjoint learning, the mean difference is $-0.60 \pm 0.66$, i.e., a best loss ratio of about $0.25\times$, with $p = 2.7 \times 10^{-3}$; against iCEM, it is $-3.20 \pm 0.79$, i.e., a ratio of about $6.3 \times 10^{-4}\times$, with $p = 5.0 \times 10^{-5}$. SCNC is therefore significantly better than both baselines at the $0.01$ level. Consistent with the sensitivity-variation analysis of Section~\ref{numerical ds_dlam}, the largest gain over adjoint learning occurs on the instability-prone tube, whereas on the smooth rod the two gradient-based methods are statistically indistinguishable and both dominate iCEM by more than two orders of magnitude.

\section{Scaling of computation time and memory with timesteps and discretization}
\label{scaling K N}

\begin{figure*}[t]
    \centering
    \includegraphics[width=1.0\textwidth]{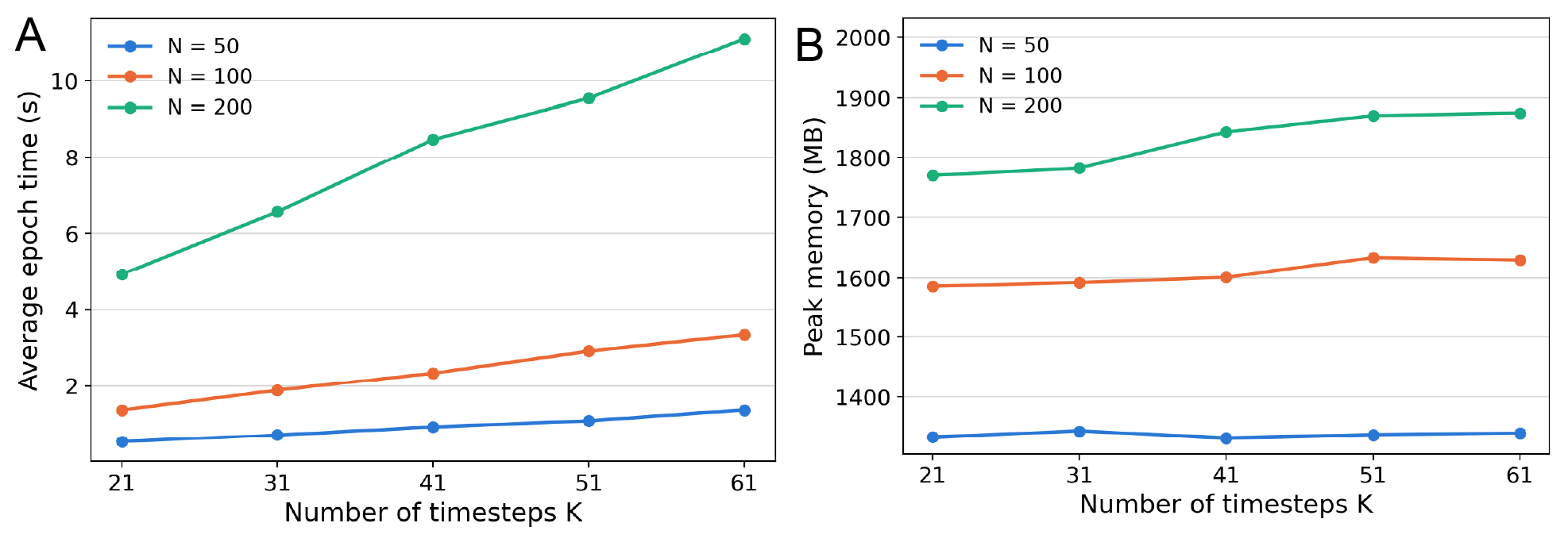}
    \caption{Computational cost of SCNC as a function of the number of timesteps $K$ for center-line discretizations with $N$ nodes. (A) Average epoch time grows approximately linearly in $K$ and superlinearly in $N$. (B) Peak memory is only affected by $N$ and is nearly independent of $K$.}
    \label{D_K}
\end{figure*}

Since the constant of Theorem~1 in the main paper contains the term $M_a C_S \Delta\lambda$ and the finite-difference stiffness derivatives carry an $\mathcal O(\Delta\lambda)$ truncation error, one may be tempted to shrink the frozen-tangent gradient error simply by refining the homotopy discretization, i.e., increasing the number of timesteps $K$. Figure~\ref{D_K} quantifies what this costs: we run SCNC on the node-targeting task with $K \in \{21, 31, 41, 51, 61\}$ and center-line discretizations of $N \in \{50, 100, 200\}$ nodes, measuring the average time per epoch and the peak memory with the same protocol as in the main paper.

\subsection{Time}
Every homotopy step performs one warm-started damped-Newton equilibrium solve plus the cached stiffness assembly of Algorithm~\ref{alg:scnc}, and the backward pass of Algorithm~\ref{alg:adjoint} performs one adjoint linear solve per step. The per-step cost is therefore independent of $K$, giving
\begin{equation}
    T
    =
    \mathcal O\!\left(K \cdot \bar I_N \cdot c(N)\right)
    =
    \mathcal O(K),
\end{equation}
for a fixed discretization, where $\bar I_N$ is the average number of Newton iterations per solve and $c(N)$ is the cost of one iteration (assembling and factorizing the tangent stiffness of $\mathcal O(N)$ DOF). Figure~\ref{D_K}(A) confirms this: for every $N$, the epoch time grows essentially linearly in $K$. The growth in $N$ is superlinear because both the factorization cost $c(N)$ and the iteration count $\bar I_N$ increase with the number of DOF.

\subsection{Memory}
In contrast, Figure~\ref{D_K}(B) shows that the peak memory is nearly flat in $K$. This is because the peak is dominated by the libraries runtime together with the working set of a single Newton solve, which depend on $N$ but not on $K$; the per-step cached stiffness matrices are small in comparison. Consequently, memory differs visibly across discretizations $N$ but varies only marginally with the number of timesteps.

\subsection{Implication}
These measurements show that refining the timestep is an expensive way to control the gradient error. Although the frozen-sensitivity error scales as $\Delta\lambda\, C_S \epsilon$ (Eq.~(19) of the main paper), doubling $K$ at best halves this term and the computation time doubles accordingly.

\bibliography{aaai2027}